\documentclass[journal,onecolumn,12pt, draftclsnofoot]{IEEEtran}
\usepackage[colorlinks=true,citecolor=blue]{hyperref}

\usepackage{epsfig}
\usepackage{svg}
\usepackage{tikz}
\usetikzlibrary{shapes,arrows.meta,positioning}
\graphicspath{{./Figures/}}
\usepackage{multirow, multicol}
\usepackage{amsmath,amsfonts}
\usepackage{comment}
\usepackage{algpseudocode}
\usepackage{orcidlink}
\usepackage{epstopdf}
\usepackage{nomencl}
\usepackage{algorithm}

\usepackage{lineno}

\usepackage{nomencl}
\usepackage{algorithm}
\algnewcommand{\Inputs}[1]{%
  \State \textbf{Inputs:} \Statex \hspace*{\algorithmicindent}\parbox[t]{.8\linewidth}{\raggedright #1}
}

\algnewcommand{\Outputs}[1]{%
  \State \textbf{Outputs:} \Statex \hspace*{\algorithmicindent}\parbox[t]{.8\linewidth}{\raggedright #1}
}
\algnewcommand{\Initialize}[1]{%
  \State \textbf{Initialize:}
  \Statex \hspace*{\algorithmicindent}\parbox[t]{.8\linewidth}{\raggedright #1}
}
\usepackage[caption=false,font=normalsize,labelfont=sf,textfont=sf]{subfig}

\makenomenclature

\newcommand\copyrighttext{%
\footnotesize This work has been submitted to the IEEE for possible publication. Copyright may be transferred without notice, after which this version may no longer be accessible.}
\newcommand\copyrightnotice{%
\begin{tikzpicture}[remember picture,overlay]
\node[anchor=south,yshift=10pt] at (current page.south) {\fbox{\parbox{\dimexpr\textwidth-\fboxsep-\fboxrule\relax}{\copyrighttext}}};
\end{tikzpicture}%
}

\begin{document}
\markboth{Adesunkanmi \MakeLowercase{\textit{et al.}}: Multi-Modal Drift Forecasting of Leeway
Objects}%
{Adesunkanmi \MakeLowercase{\textit{et al.}}: Multi-Modal Drift Forecasting of Leeway Objects}
\title{ Multi-Modal Drift Forecasting of Leeway Objects via Navier-Stokes-Guided CNN and Sequence-to-Sequence Attention-Based Models}

\author{Rahmat K. Adesunkanmi\orcidlink{0000-0002-5483-5076}, Alexander W. Brandt\orcidlink{0000-0000-0000-0000},  Masoud Deylami\orcidlink{0009-0006-7136-4583}, Gustavo A. Giraldo Echeverri\orcidlink{0009-0000-4716-8554}, Hamidreza Karbasian\orcidlink{0000-0000-0000-0000}, Adel Alaeddini\orcidlink{0000-0000-0000-0000}
\thanks{Rahmat K. Adesunkanmi is with the Departments of Mechanical and Electrical Engineering, Southern Methodist University, Dallas, TX, 75205, USA (email: radesunkanmi@smu.edu)  \\
Alexander W. Brandt,  Masoud Deylami, Gustavo A. Giraldo Echeverri, Hamidreza Karbasian, Adel Alaeddini are with the Department of Mechanical Engineering, Southern Methodist University, Dallas, TX, 75205, USA
(emails: awbrandt@smu.edu, mdeylami@smu.edu, ggiraldoecheverri@smu.edu, hkarbasian@smu.edu, aalaeddini@smu.edu) }
\thanks{(Corresponding author: Adel Alaeddini aalaeddini@smu.edu)}}

\maketitle
\copyrightnotice
\begin{abstract}
Accurately predicting the drift (displacement) of leeway objects in maritime environments remains a critical challenge, particularly in time-sensitive scenarios such as search and rescue operations. In this study, we propose a multi-modal machine learning framework that integrates Sentence Transformer embeddings with attention-based sequence-to-sequence architectures to predict the drift of leeway objects in water. We begin by experimentally collecting environmental and physical data, including water current and wind velocities, object mass, and surface area, for five distinct leeway objects. Using simulated data from a Navier-Stokes-based model to train a convolutional neural network on geometrical image representations, we estimate drag and lift coefficients of the leeway objects. These coefficients are then used to derive the net forces responsible for driving the objects' motion. The resulting time series, comprising physical forces, environmental velocities, and object-specific features, combined with textual descriptions encoded via a language model, are inputs to attention-based sequence-to-sequence long-short-term memory and Transformer models, to predict future drift trajectories. We evaluate the framework across multiple time horizons ($1$, $3$, $5$, and $10$ seconds) and assess its generalization across different objects. We compare our approach against a fitted physics-based model and traditional machine learning methods, including recurrent neural networks and temporal convolutional neural networks. Our results show that these multi-modal models perform comparably to traditional models while also enabling longer-term forecasting in place of single-step prediction. Overall, our findings demonstrate the ability of a multi-modal modeling strategy to provide accurate and adaptable predictions of leeway object drift in dynamic maritime conditions.
\end{abstract}


\section{Introduction}
\begin{figure}[t]
\centering
\includegraphics[width  = 1\linewidth]{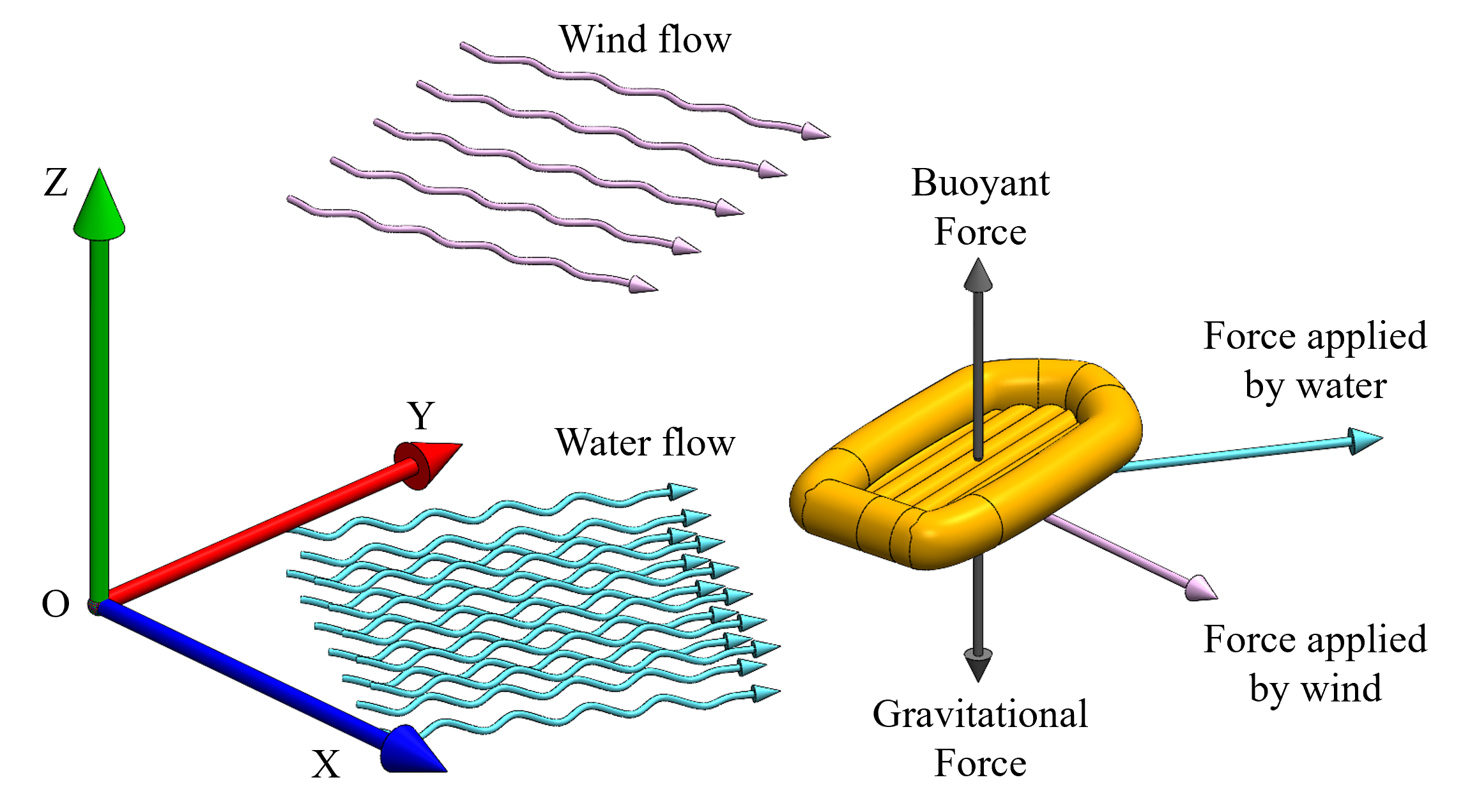}
 \vspace{-0.3in}
\caption{Water and wind-induced forces acting on a floating object: upward buoyancy balances gravity, while horizontal forces (wind drag and water drag) contribute to net drift. Lift, which is opposing gravity and perpendicular to the direction of the fluid flow, and drag, which acts parallel to it.}
\label{fig:force_diagram}
\end{figure}

 \IEEEPARstart{I}{n} recent years, methods to improve the efficiency of Search and Rescue (SAR) operations in maritime environments have gained popularity~\cite{kwesi2019scenario,akyuz2017marine}. The drift characteristics of objects play a critical role in enabling rapid and effective SAR operations~\cite{yang2021novel} due to the complex and dynamic nature of sea conditions. These predictions are used to narrow down search areas for floating devices (such as life rafts, debris, inflatables, or persons-in-water) to increase the chances of successful operations and reduce response times~\cite{kim2023validation,NI20101169,jmse12020357}. Modeling the trajectory of a leeway object on the ocean surface lies in the empirical determination of coefficients that relate the combined aerodynamic and hydrodynamic forces to object motion. As illustrated in Figure~\ref{fig:force_diagram}, the gravitational force pulls the object downward, while the buoyant force acts upward, opposing gravity and determining the submerged level of an object. When a fluid moves around an object, it exerts a force on the object, which can be broken into two components: lift, which is commonly associated with an upward force opposing gravity and perpendicular to the direction of the fluid flow, and drag, which acts parallel to it. Aerodynamic drag, one of the dominant forces,  acts in the direction of the wind and depends on the object's exposed area above the water and the relative wind speed. Hydrodynamic drag arises from interaction with water currents and depends on the submerged area, shape, and current velocity~\cite{HowWind}. These forces combine to determine the net motion of the object. Traditionally, they are grounded mainly in physics-based models that parameterize object drift as a linear function of wind speed, with distinct coefficients for downwind and crosswind components~\cite{article10,  10.3389/fmars.2022.1017042,WU2023113444}. Despite its importance, the general task of predicting the motion of objects in water remains a central challenge due to the dynamic and uncertain nature of oceanographic conditions~\cite{BREIVIK2011100}. 
 
Many works like \cite{breivik2008operational, kang1999field, allen1997leeway} have emerged over the years to help understand the definition, modeling, and prediction of drift. Numerous studies incorporate wind, current, and wave field measurements, either as direct model inputs or for the estimation of leeway coefficients and other target-specific parameters~\cite{breivik2008operational, Bhaganagar,1532757, articleLi, TU2021110134}. The Leeway model, initially developed by the U.S. Coast Guard and later implemented in various computational frameworks, including OpenDrift~\cite{dagestad2018opendrift,rohrs2018effect,su1997predicting}, is a typical example. The models, which have been proven effective under specific conditions,  capture the relationship between feature characteristics through parametric equations. However, performance can decrease in the presence of sparse or noisy environmental inputs and require manual calibration of object-dependent parameters.
Furthermore, such models generally assume static relationships and do not adapt to evolving environmental conditions~\cite{ev,sutherland2020evaluating}. Artificial Intelligence (AI) and Machine Learning (ML) offer new opportunities to overcome the limitations of static models by leveraging temporal relationships amongst observational features, incorporating complex patterns in environmental forcing, adapting to varying conditions, as well as uncovering hidden relationships. In this study, we address these limitations by proposing a combined experimental and multi-modal ML framework for predicting the drift of leeway objects. The key contributions of this article are summarized as follows.

\begin{enumerate}
\item We sourced data such as the water and wind velocities, objects' masses and surface areas, through controlled experiments on five distinct boats. 

\item We developed a Navier-Stokes-based simulation model to estimate the drag and lift coefficients of synthetic geometrical image data, which serves as input to train a Convolutional Neural Network (CNN) model. This trained model is used to compute drag and lift coefficients acting on each leeway object. We calculated the objects' time-wise forces by combining physical and environmental parameters.

 \item We encoded boat-specific textual descriptions using a pre-trained Language Model (LM) built upon the Sentence Transformer architecture and utilized them alongside temporal, physical, and environmental data. This model is called the sentence-BERT introduced in \cite{reimers-2019-sentence-bert}. 

\item  We present attention-based sequence-to-sequence frameworks to forecast the drift of leeway objects, utilizing both temporal data and textual data to achieve multi-modal time series forecasting. 

 \item  The language-model-inspired architecture is used to forecast drift trajectories over multiple time horizons in seconds($1s$, $3s$, $5s$, and $10s$), showing robustness in short- and near-term sampling.
\end{enumerate} 
 
 We demonstrate the feasibility of multi-modal modeling in modern deep learning for real-world maritime applications. The rest of this paper is organized as follows: Some related works are discussed in Subsection~\ref{Sec:litrev}. In Section~\ref{Sec:Method}, we discuss the proposed model, detailing the experimental setup, force modeling approach, multi-modal drift prediction framework, and evaluation strategy. In Section~\ref{Sec:Results}, we validate the model using experimental data, show the results of the predictions, and perform some ablation studies. We discuss our findings and limitations in Section~\ref{Sec:Discussions}. Finally, our conclusions are summarized in Section~\ref{Sec:Conclusions}.

\subsection{Related Works}\label{Sec:litrev}
Current research has demonstrated the effectiveness of ML architectures for encapsulating complexities in maritime data~\cite{articleLi}. These advancements have been validated in numerous operational settings, including field deployments in Galveston Bay~\cite{Bhaganagar}, the Korea Strait~\cite{KSCOE.2022.34.1.11}, South China Sea~\cite{TU2021110134} and custom open-sea trials~\cite{jmse12122262}. Some studies employ real-world datasets, such as GPS-tracked surface drifters from programs like the Defense Advanced Research Projects Agency (DARPA) Ocean of Things~\cite{8867398}, field deployments of mannequins and surface buoys~\cite{jmse12122262, 9705735,article2}, or life rafts~\cite{TU2022112158}. Others supplement with synthetic trajectories generated from ocean circulation models~\cite{jenkins2023}, enabling assessment of generalization to unseen conditions. Several studies enhance ML models by augmenting limited observational datasets with synthetic trajectories derived from realistic ocean model simulations~\cite{article2,jenkins2023,app10228123,6848745}.  

Architectures ranging from Deep Neural Networks (DNNs), Long Short-Term Memory (LSTM), Gated Recurrent Unit (GRU), CNN, Bidirectional GRU (BiGRU), and attention mechanism (AM)(CNN-BiGRU-Attention), and hybrid feedforward-recurrent models, have outperformed or improved classical drift models~\cite{articleLi,article2,jmse12060958,jmse12020323} such as the Regional Ocean Modeling System (ROMS)~\cite{SHCHEPETKIN2005347} and Modular, Object-Oriented Hydrodynamic Modeling System (MOHID)~\cite{miranda2000mohid}, showing consistent error reductions~\cite{app10228123,jmse12111933,10489575}.. When trained and evaluated on large, real-world drifter datasets, DNNs have achieved a significant error reduction in short-term forecasting compared to traditional physics-based models~\cite{9705735,2022EA002362}. Fully Connected Neural Networks (FCNNs) have been able to adjust predictions and improve accuracy in operational SAR environments by assimilating ground-truth data in real-time and have been effective in a dynamic correction during SAR operations~\cite{jmse12122262,1532757}. Frameworks have also incorporated spatial convolutions, attention mechanisms, and multi-scale feature fusion to capture complex environmental influences better~\cite{jmse12060958,jmse12020323}. Gaussian Process Regression (GPR) was used to predict ship motion trajectories by modeling speed, sway, surge, and drift angles under varying hydro-meteorological conditions. This approach achieved interpretable results but struggled with nonlinear oceanic dynamics~\cite{ZHANG2023114905}. LSTM networks are particularly suitable for drift prediction due to their ability to model sequential data with temporal dependencies~\cite{articleLi, Mlaskamit}. The LSTM-DNN hybrid model is used to refine predictions using multi-modal inputs by leveraging the temporal modeling capabilities of LSTM while incorporating the nonlinear mapping capabilities of dense layers to capture complex relationships between environmental variables and drift behavior~\cite{articleLi,jmse12061016}. 

Most ML-based drift prediction frameworks rely solely on numerical sensor data and do not often incorporate unstructured information, such as text-based metadata. Parallel to this, the rapid advancement of Large Language Models (LLMs) and Transformer architectures has expanded the frontier of sequential data modeling. More recently, Transformer-based models~\cite{vaswani2023attentionneed} have demonstrated strong performance in trajectory prediction applications~\cite{GAO2024111744}. LLMs have shown success in time series forecasting~\cite{zhang2024large,gruver2024largelanguagemodelszeroshot}, event prediction~\cite{shi2023language}, and multi-modal learning~\cite{wu2023multimodal}.  Transformer models' attention mechanisms allow encourages more effective long-range dependencies than recurrent networks and adapt to both structured and unstructured input data, making them great for applications involving complex, heterogeneous datasets. Recently, in maritime applications, the double-branch adaptive span attention (DBASformer) model captures both cross-time and cross-dimension relationships in drift trajectory sequences~\cite{jmse12061016} and addresses the challenge of variable-length sequences and complex inter-dependencies between multiple environmental variables that influence drift behavior~\cite{jmse12061016}. Multi-functional buoy drift prediction has benefited from the ResNet-GRU-Attention (RGA) model, which addresses nonlinear time series drift prediction problems~\cite{s22145120}. The incorporation of external knowledge and natural language paraphrases has been shown to positively affect the predictive performance of LLMs for time series applications~\cite{tang2024tsfllms}. Time-LLM represents a framework approach that involves reprogramming input time series with text prototypes before feeding them into frozen LLMs to align temporal and linguistic modalities~\cite{jin2024timellmts}. 
The self-attention mechanism enables ML models to learn dependencies between all pairs of time steps, regardless of their distance in the sequence. The attention helps a model focus on the most relevant parts of the input when making a prediction or generating an output. It does this by computing weighted combinations of input elements, assigning higher weights (more "attention") to the parts that are more important for the task, and is mathematically defined through a series of linear transformations and attention computations: the query, key, and value projections. Attention scores measure how much attention each query should pay to each key, and the Softmax converts these scores into probabilities. The integration of attention mechanisms in time series forecasting and the utilization of pretrained Sentence Transformers for text conversion are two powerful applications of modern deep learning. Sentence Transformers provide extensive support for text encoding using tokenizers with pretrained models to encode texts into embeddings, following different strategies depending on the architecture.
\section{Framework I: Experimental, Calculated and Simulated Data Collection}\label{Sec:Method}
This study presents a multi-modal ML framework for forecasting the future trajectory of drifting leeway objects. The objective is to predict the future trajectories over a window of length $\ell_d$, denoted by $\mathcal{Y}_{t+1:t+\ell_d}= \{d^x_{t+i}, d^y_{t=i}\}_{i=1}^{\ell_d}$, given past input data of window length $\ell_e$, denoted by $\mathcal{X}_{t-\ell_e+1:t} = \{\mathcal{X}_{t-j}\}_{j=\ell_e-1}^{0}$ using multi-modal attention-based sequence-to-sequence models. The dataset is sampled at fixed time horizons $t_h$. Since the original time series is down-sampled by a factor of $t_h$, the effective temporal length $T$ of the dataset becomes $T= \frac{T_\mathrm{original}}{t_h}$. The input data includes environmental variables (e.g., wind and current velocities), physical attributes of the objects, and textual descriptions. A multi-modal attention-based sequence-to-sequence model is employed to learn the mapping between past and future drift behavior under varying sampling intervals. 

The first framework is structured into two key stages: (i) experimental collection of key environmental and physical parameters for five distinct leeway objects in water; (ii) estimation of drag and lift coefficients via a CNN trained on geometrical image data with labels derived from a Navier-Stokes-based simulation model.

\subsection{Experimental Setup and Data Collection}\label{sec:experiments}
\begin{figure}[!ht]
    \centering
    \includegraphics[width=1\linewidth]{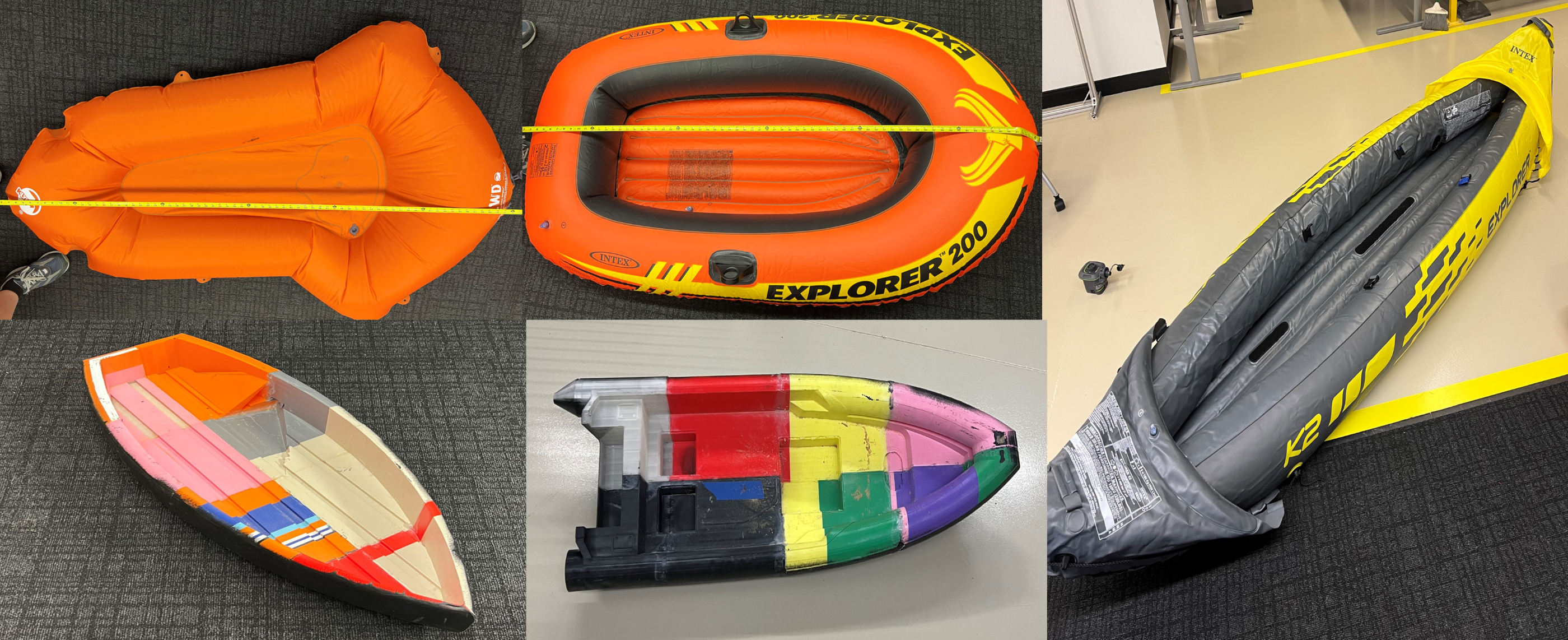}
     \vspace{-0.3in}
    \caption{The five leeway objects tested in the experiments. Top left is the deformed inflatable, top middle is the orange inflatable, the far right boat is the banana boat, the bottom left is the mainly orange 3-D printed boat, and the bottom middle is the mainly red and black 3-D printed boat. The surface areas of the objects exposed to wind from any direction, $A_a$ are calculated based on a 3-D computer-aided design (CAD) model of each object in \texttt{SolidWorks}, $A_a=[4.067, 2.759, 5.314, 0.834 0.932]\,m^2$ and masses $m_o = [1, 1.900,5.634, 4.022, 5.000 ]\,kg$ respectively. The areas submerged in water $A_w$ are assumed to be $10\%$ of $A_a$ for inflatables and $25\%$ of $A_a$ for the 3-D printed boats.} 
    \label{fig:leewayobj}
\end{figure}

\begin{figure*}[t]
    \centering
    \includegraphics[width=1\linewidth, height = 0.3\linewidth]{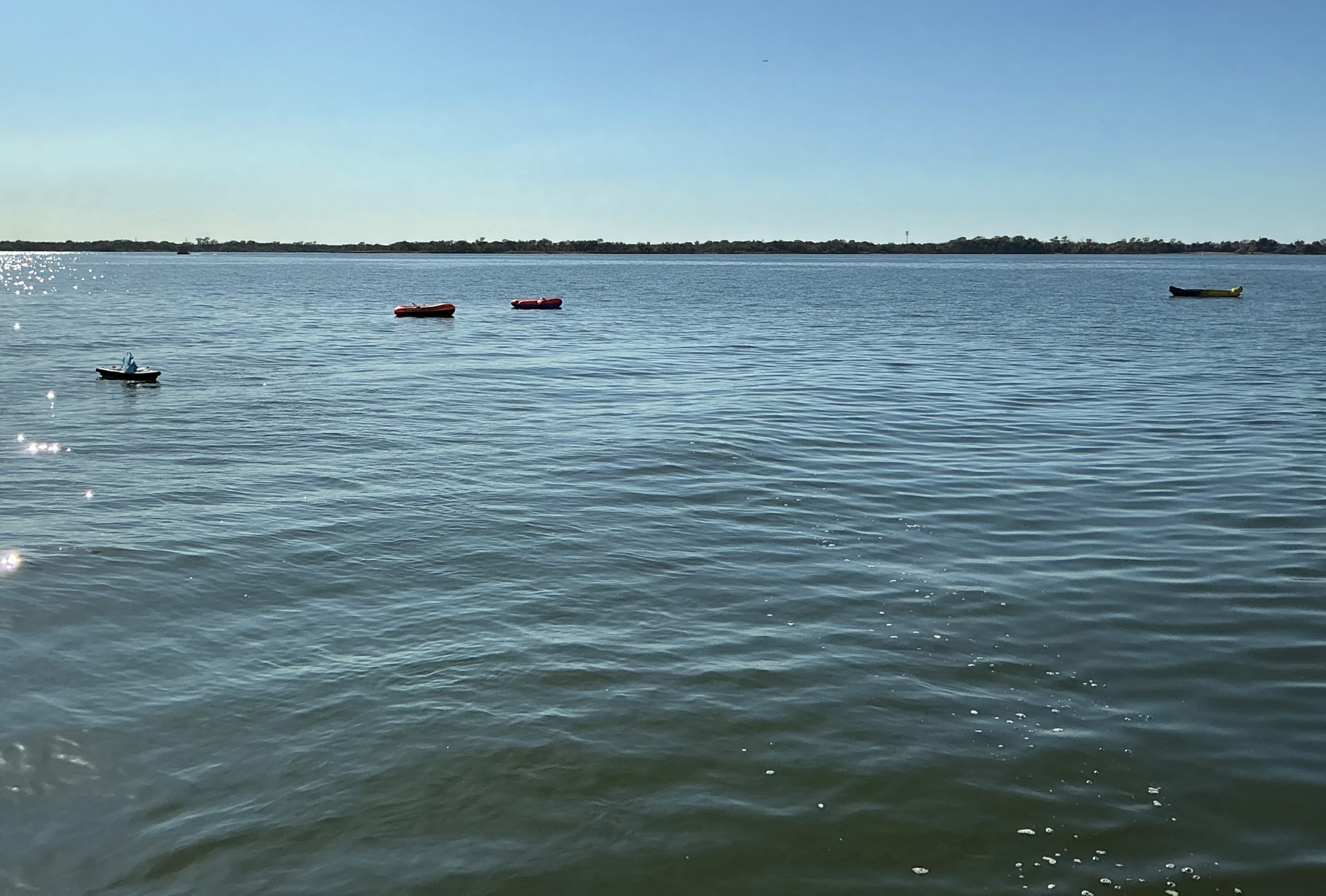}
     \vspace{-0.3in}
    \caption{Leeway objects docked in water at {\it Lewisville Lake in Little Elm, Texas} with coordinates $33.1130767^\circ$ North, $96.9735092^\circ$ West. The water current velocities are within $0.2{-}0.4 \, \mathrm{ft/s} \approx 0.0610{-}0.1219 \, \mathrm{m/s}$, and the wind velocities are within $2{-}5\,\mathrm{mph} \approx 0.8941{-}2.2352\,\mathrm{m/s}$.}
    \label{fig:leewaywater}
\end{figure*}

Experimentally collating data involved monitoring the drifting position $\mathcal{Y}$ of the five leeway objects in water, as well as the relevant environmental velocities: wind ($v_a$) and current ($v_w$), for a specific amount of time $T$. These tested leeway objects are illustrated in Figure~\ref{fig:leewayobj}. Three of the objects are inflatables, while the other two are $3$-dimensional (3-D) printed boats. 
 The boats were docked on {\it Lewisville Lake in Little Elm, Texas}, at $33.1130767^\circ N$, $96.9735092^\circ W$ as depicted in Figure~\ref{fig:leewaywater}. 

The wind speed and direction were measured using a {\it Young USA 05103V wind monitor}\footnote{https://www.youngusa.com/product/wind-monitor/} with $0-5\,V$ direct current (VDC) output affixed to a tripod at a height of $2.066\,\mathrm{m}$ above the waterline. Attached to the wind monitor was an Arduino Micro that converts the analog signal to a digital one. The Arduino was connected to a $5V$ portable power supply and a Raspberry Pi Zero W, which logged the data to a $64GB$ secure digital (SD) card and could be accessed via a secure shell (SSH) on a laptop computer. This feature proved to be important so that the wind monitor voltage output could be calibrated against magnetic north, as well as allowing for periodic verification that the logging was being completed as expected. To allow for SSH connections, the Raspberry Pi Zero W was connected to a Wi-Fi network. The wind data was logged at a frequency of $1\,\mathrm{Hz}$. The wind speed and direction relative were measured relative to true north at a height of $2.0066\,\mathrm{m}$ above the waterline and was then converted to the $10\,\mathrm{m}$ wind speed in the local coordinates corresponding to magnetic East and North, respectively using the power-law wind profile: 
\begin{align}
   v_a =v_\mathrm{ref}\left(\frac{z}{z_\mathrm{ref}}\right)^\beta, 
  \label{eqn:10m}
\end{align}

\noindent where $v_a$ is the wind speed in $\mathrm{m/s}$ at reference height \noindent where $v_a$ is the wind speed in $\mathrm{m/s}$ at reference height $z=10\,\mathrm{m}$, $v_\mathrm{ref}$ is the known wind speed at reference height $z_\mathrm{ref}$, and $\beta$ is a constant, determined experimentally. $\beta=0.10$ for open water with near-neutral stability conditions~\cite{hsu1994}.

The water current speed was monitored using a {\it Global Water Flow Probe Digital Water Velocity Meter}\footnote{https://www.fondriest.com/global-water-flow-probes.htm} with a $5.5 - 15$ feet telescoping pole. Attached to the pole was a flexible ribbon that served as a windsock that moved with the fluid flow to indicate the direction of the currents. From that, the direction of the currents was read against magnetic north using a portable cellular device. Due to the digital readout being the most widely used type of readout for flow probes of this type, the readings were conducted every $5$ minutes throughout the experiment by submerging the flow probe into the water about $2\,\mathrm{m}$ and rotating the sensor until the impeller spun at its maximum value for a given physical location. If no readings could be detected, the device was placed on the opposite side of the pontoon boat, such as to measure the currents without interference from the boat's wake behind the flow.

\begin{figure}[!h]
     \centering
     \includegraphics[width=1\linewidth]{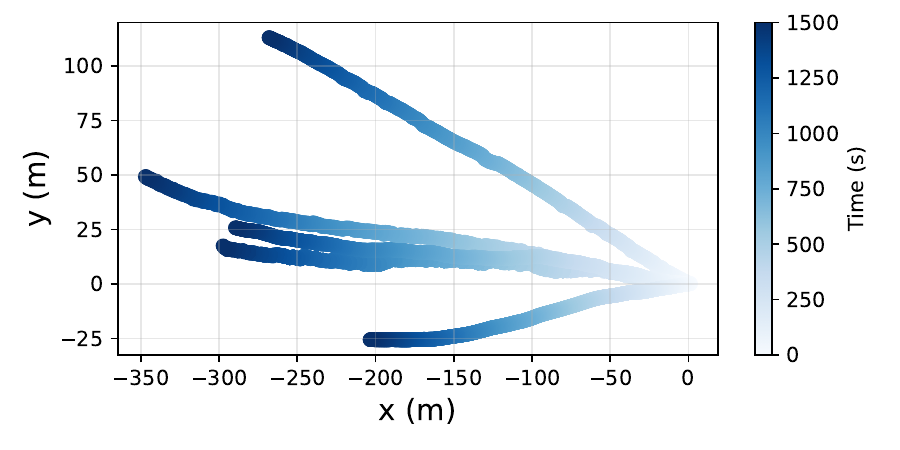}
      \vspace{-0.3in}
     \caption{The drift trajectories of leeway objects in the local coordinate system: eastward $(x)$ and northward $(y)$ directions in meters with respect to time in seconds}
     \label{fig:drift}
 \end{figure}

The components in the local coordinate system, illustrated by scatter plots in Figure~\ref{fig:drift}, are determined from trigonometric identities, given the magnitude and direction of the wind. 
To convert the longitude and latitude coordinates of the drift trajectories from the GPS measurements into Cartesian coordinate readings, the \texttt{MATLAB} function \texttt{wgs84Ellipsoid} was utilized to create an instance of  \texttt{referenceEllipsoid} for the World Geodetic System of 1984 (WGS84) reference ellipsoid. This object, along with a reference height of the lake from the reading closest to the experiment date, taken by the Texas Water Development Board,  was inputted into the \texttt{geodetic2enu} function to output the $\mathcal{Y}$ coordinate equivalent. 

\subsection{Objects Drag and Lift Coefficients Estimation and Prediction via Navier-Stokes-Guided CNN }
At this stage, our objective is to estimate the drag and lift coefficients that are critical for modeling the leeway objects' motion in water. We adopt a two-step approach by first generating a comprehensive dataset of geometrical objects with known coefficients using Computational Fluid Dynamics (CFD) simulations and second, training a CNN to learn the mapping from object geometry to these coefficients.

\subsubsection{Dataset for estimation of Drag and Lift Coefficients using simulation}\label{sec:drag}
\begin{figure}[!h]
    \centering \includegraphics[width=1\linewidth]{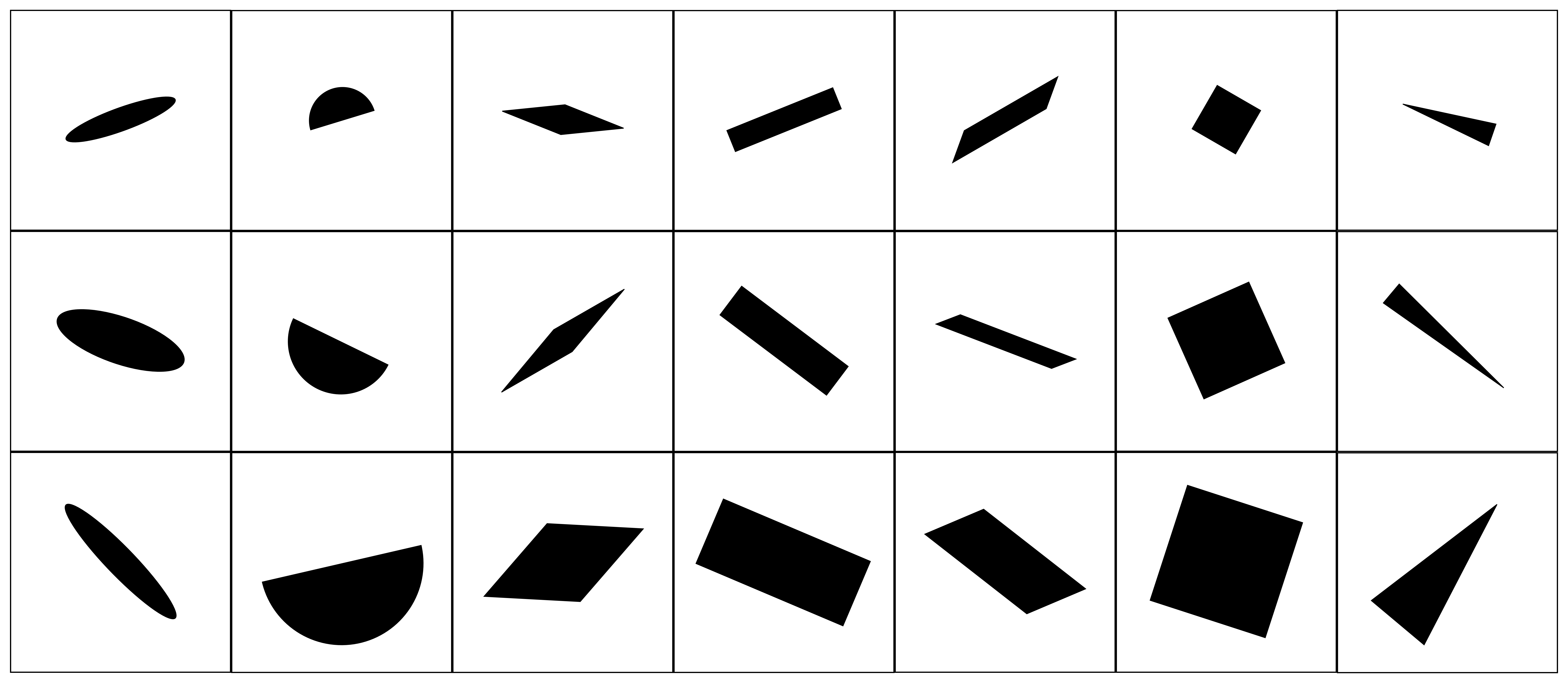}
     \vspace{-0.3in}
    \caption{Demonstration of geometries used in CFD simulations.}
    \label{fig:Geometries}
\end{figure}

We introduce an automated CFD framework based on the Navier-Stokes equation that computes drag and lift coefficients for a diverse set of 2-D object geometries, varying their size and orientation relative to the flow direction to enhance variability and ensure broad, randomized coverage of the feature space. The resulting simulation-informed coefficients serve as ground-truth labels for the CNN. As shown in Figure~\ref{fig:Geometries}, the geometries are a diverse set of commonly seen 2-D object shapes: ellipses, half-circles, rhombi, rectangles, squares, and triangles. The rationale behind the selection of these simple primitives stems from the observation that complex geometries can often be decomposed into or approximated by the superposition and transformation of basic ones.

\begin{figure}[ht]
    \centering
    \begin{tikzpicture}
        \node[inner sep=0pt] at (0,0) {\includegraphics[width=.9\linewidth]{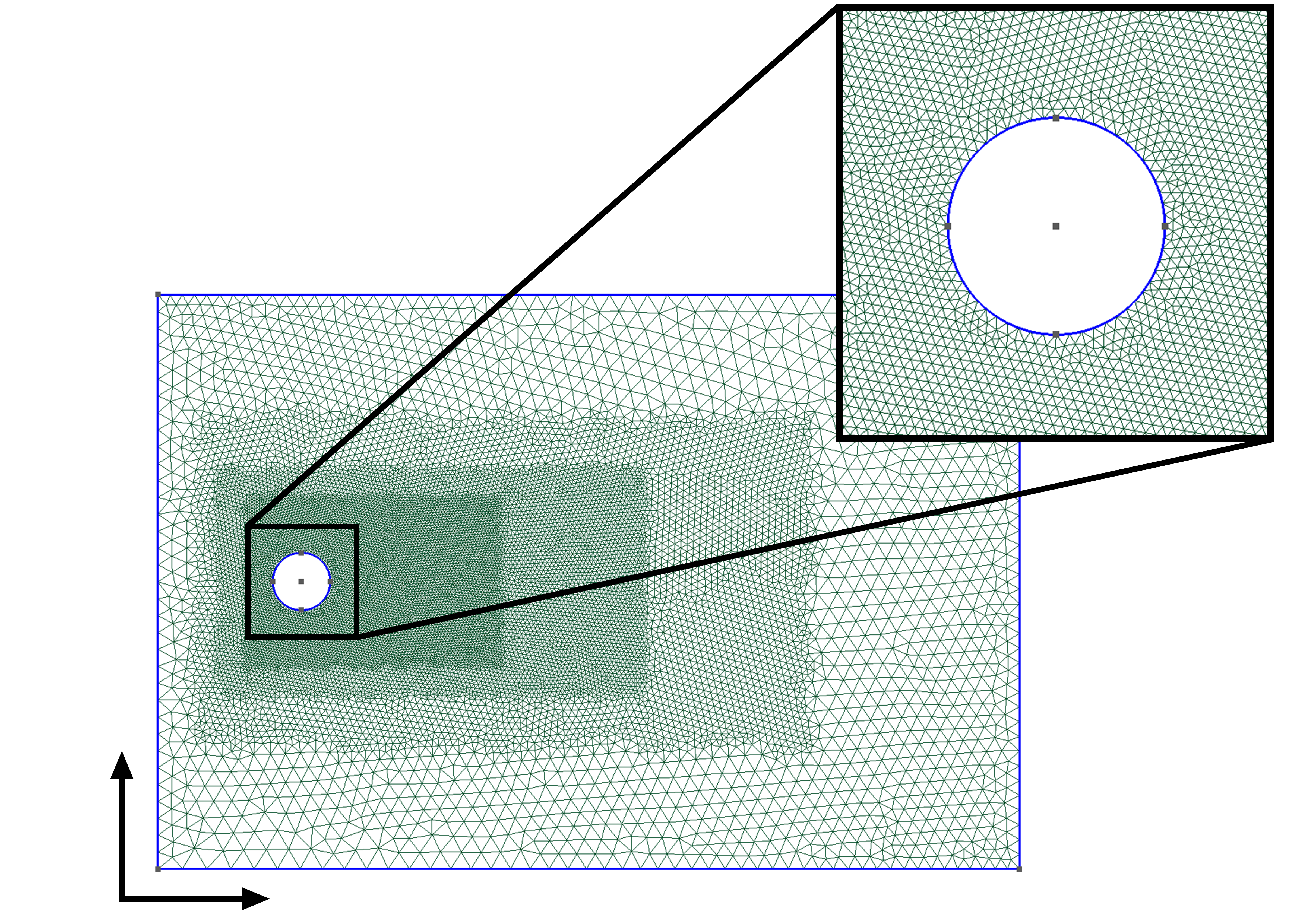}};
        
        \node at (-2.4,-3.0) {\textbf{$x$}};
        \node at (-3.6,-1.8) {\textbf{$y$}};
        
        \node at (-3.95,-0.1) {\small $u = u_\infty$};
        \node at (-4.1,-0.5) {\small $v = 0$};
        \node at (-4.0,-1.0) {\small $\frac{\partial p}{\partial n} = 0$};
        
        \node at (3.4,-1.0) {\small $p = p_\infty$};
        \node at (3.3,-1.45) {\small $\frac{\partial u}{\partial n} = 0$};
        \node at (3.3,-2.0) {\small $\frac{\partial v}{\partial n} = 0$};
        
        \node[fill=white, inner sep=1pt] at (2.8,2.6) {\small $u,v = 0$};
    \end{tikzpicture}
     \vspace{-0.3in}
    \caption{Computational domain, boundary conditions, and multi-block unstructured mesh for a circular cylinder.}
    \label{fig:meshanddomain}
\end{figure}
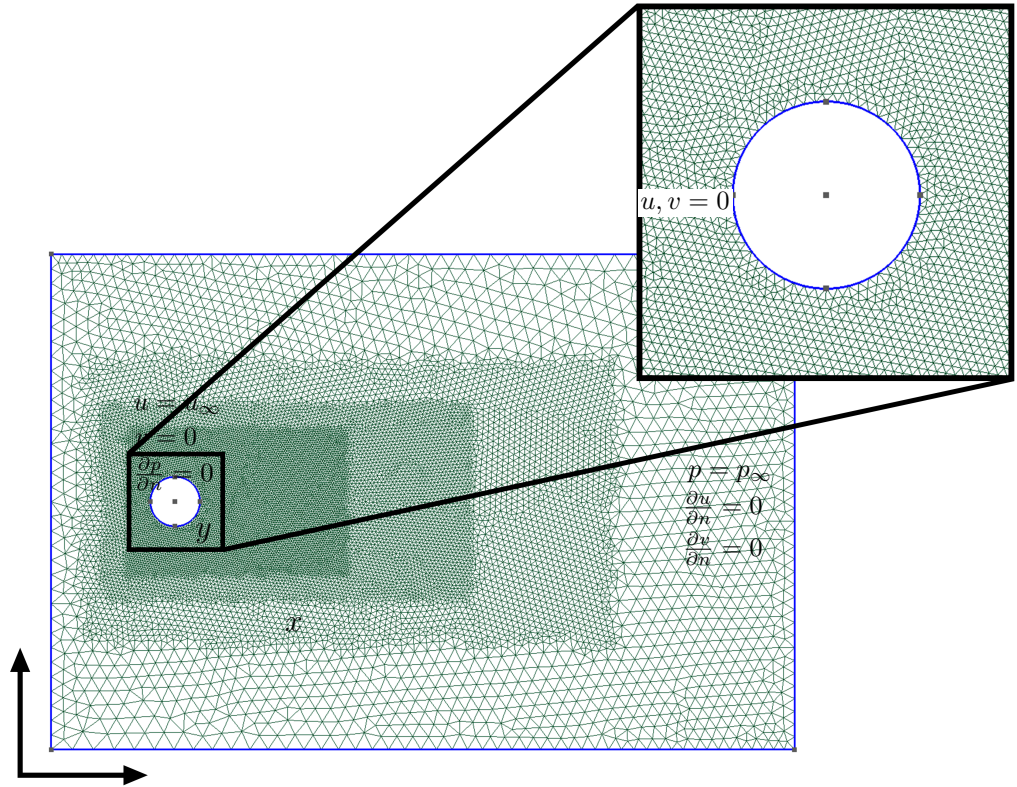

In this study, the compressible Navier-Stokes equations~\cite{Pereira2020Spectral} 
\begin{align}
\frac{\partial \mathbf{u}(\mathbf{x}, t)}{\partial t} 
+ \nabla \cdot \left( \mathbf{F}(\mathbf{u}) - \mathbf{F}_\nu(\mathbf{u}, \nabla \mathbf{u}) \right) = 0
\end{align}
are solved for laminar flow past a 2-D object using a Finite Volume (FV) approach with second-order accuracy, where $\mathbf{x} \in \mathbb{R}^2$: spatial coordinates in 2-D space,  $t$ is time, $\mathbf{u}(\mathbf{x}, t)$ is the state vector containing the conserved quantities, $\mathbf{F}(\mathbf{u})$ is the convective flux and $\mathbf{F}_\nu(\mathbf{u}, \nabla \mathbf{u})$ is the viscous flux. The time integration is performed using a third-order Runge-Kutta scheme. A 2-D computational domain extends at least $10$ times the characteristic length of the object from its boundary. It is discretized using a multi-block unstructured mesh, as shown in Figure~\ref{fig:meshanddomain}. 

The mesh is progressively refined near the object surface to capture the viscous boundary layer region accurately. This meshing strategy eliminates the need to generate a separate structured mesh for each geometry and enables automation of the framework. The upstream and lateral edges of the domain are specified using a standard velocity inlet boundary condition with $u = u_\infty$ and $v =0$, where $u$ and $v$ are velocity components, and $u_{\infty}$ is the free stream velocity. Additionally, the downstream boundary is specified using the ambient pressure $p =p_\infty$ (i.e., atmospheric pressure), and a no-slip adiabatic boundary condition is specified on the surface of the object. The time-step is set to $\Delta t = 0.05t^*$, where $t^*$ is a convective time ($t^* = \frac{l_c}{u_\infty}$, where $l_c$ is characteristic length). Note that the characteristic length for the circular cylinder is set to $l_c = D$, where $D$ is the diameter of the cylinder. 

\begin{figure}[!ht]
    \centering    \includegraphics[width=.85\linewidth, trim = 0cm 2.5cm 0cm 0cm, clip]{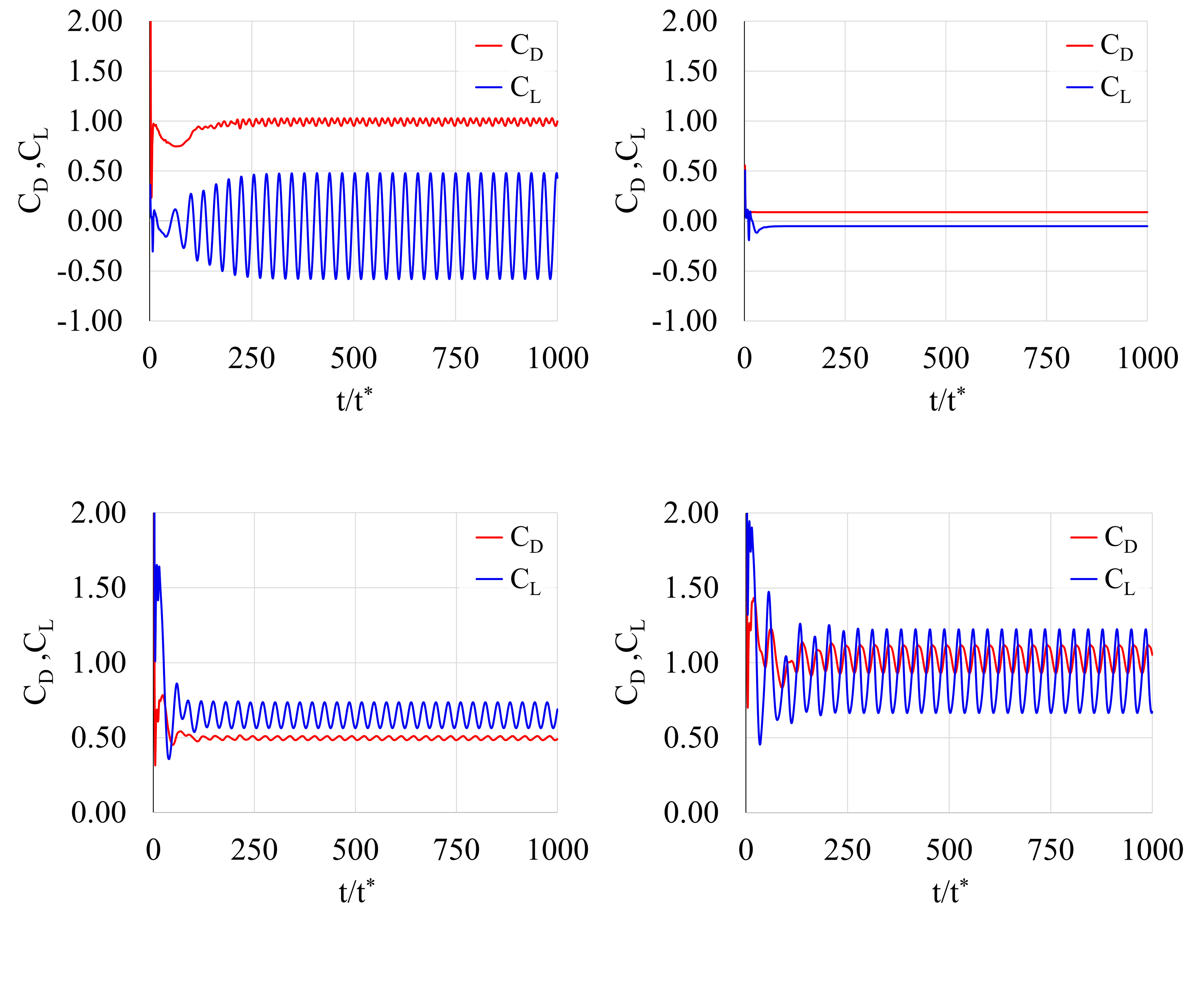}  
     \vspace{-0.3in}
    \caption{Instantaneous drag and lift coefficients for (Top left) a circular cylinder, (Top right) an elliptical cylinder oriented parallel to the flow, (Bottom left) an elliptical cylinder at $\alpha = 30^{\circ}$, and (Bottom right) an elliptical cylinder at $\alpha = 40^{\circ}$ with respect to the flow direction.}
    \label{fig:CdCl}
\end{figure}

The \texttt{Julia} dynamic programming language (version \texttt{1.8.0})~\cite{bezanson2012julia} was employed for numerical computations. Geometry and mesh generation were carried out using \texttt{GMSH} (version \texttt{4.13.1})~\cite{geuzaine2009gmsh}, and simulation results were visualized with \texttt{ParaView} (version \texttt{5.13.1})~\cite{ayachit2015paraview}. A \texttt{Python}-based script was developed to automatically generate 2-D geometries and meshes in \texttt{GMSH}, save the geometries as images, call the \texttt{Julia} solver for simulations, and store the simulation results sequentially. A total of $179$ geometries were simulated using this framework. A sufficient time-averaging period is employed to ensure accurate estimation of the aerodynamic forces acting on the object. Figure~\ref{fig:CdCl} illustrates the effect of the shape and orientation of the object on the instantaneous lift and drag coefficients. 

\begin{figure}[!ht]
    \centering
{\includegraphics[width=.6\linewidth]{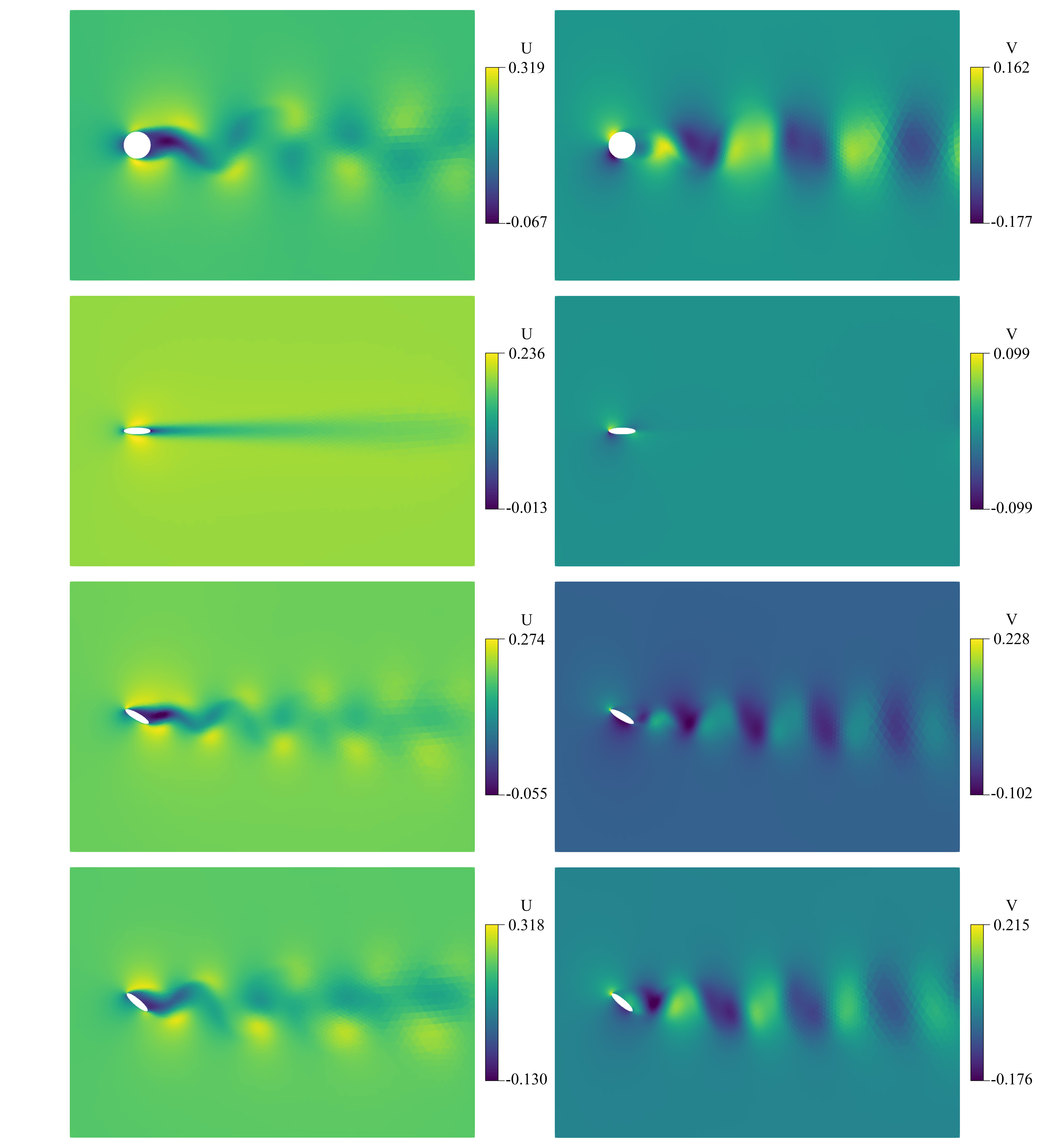}};
     \vspace{-0.3in}
    \caption{Aerodynamic loads  for flow past (from top (a) to bottom (d)) (a) a circular cylinder, (b) an elliptical cylinder oriented parallel to the flow, (c) an elliptical cylinder at $\alpha = 30^{\circ}$, and (d) an elliptical cylinder at $\alpha = 40^{\circ}$ with respect to the flow direction.}
    \label{fig:simulations}
\end{figure}

Simulation computes velocity and pressure, from which the instantaneous drag and lift forces, $F_x$ and $F_y$, on the object surface are determined. These forces are then used to calculate the corresponding instantaneous lift and drag coefficients, ${C}_L$ and ${C}_D$. Finally the time-averaged lift and drag coefficients, $\bar{C}_L$ and $\bar{C}_D$, are computed by taking average of the instantaneous values between $\frac{t}{t^*} = 500$ and $\frac{t}{t^*} = 1000$. Instantaneous coefficients are computed using the following equations:
\begin{align}
 {C}_L=\frac{{F}_y}{0.5\rho_\infty cu^2_\infty}   \quad  {C}_D=\frac{{F}_x}{0.5\rho_\infty cu^2_\infty} 
\end{align} 

\noindent where $c$ is the characteristic length of the object, and $\rho_\infty$ represents the freestream density. As illustrated in Figure~\ref{fig:simulations}, the framework is initially designed and validated for a circular cylinder across a range of Reynolds numbers, $Re =20-250$, with a constant free stream Mach number of $0.1$. Simulation results at $Re =150$ were obtained and compared with existing literature, showing good agreement~\cite{CAGNONE2013324, INOUE_HATAKEYAMA_2002}. Following this validation, the same framework was applied to other geometries and their variations.

\subsubsection{Leeway Objects' Drag and Lift Coefficients Prediction using Convolutional Neural Network}
\begin{figure}[ht]
    \centering
    \includegraphics[width = \linewidth]{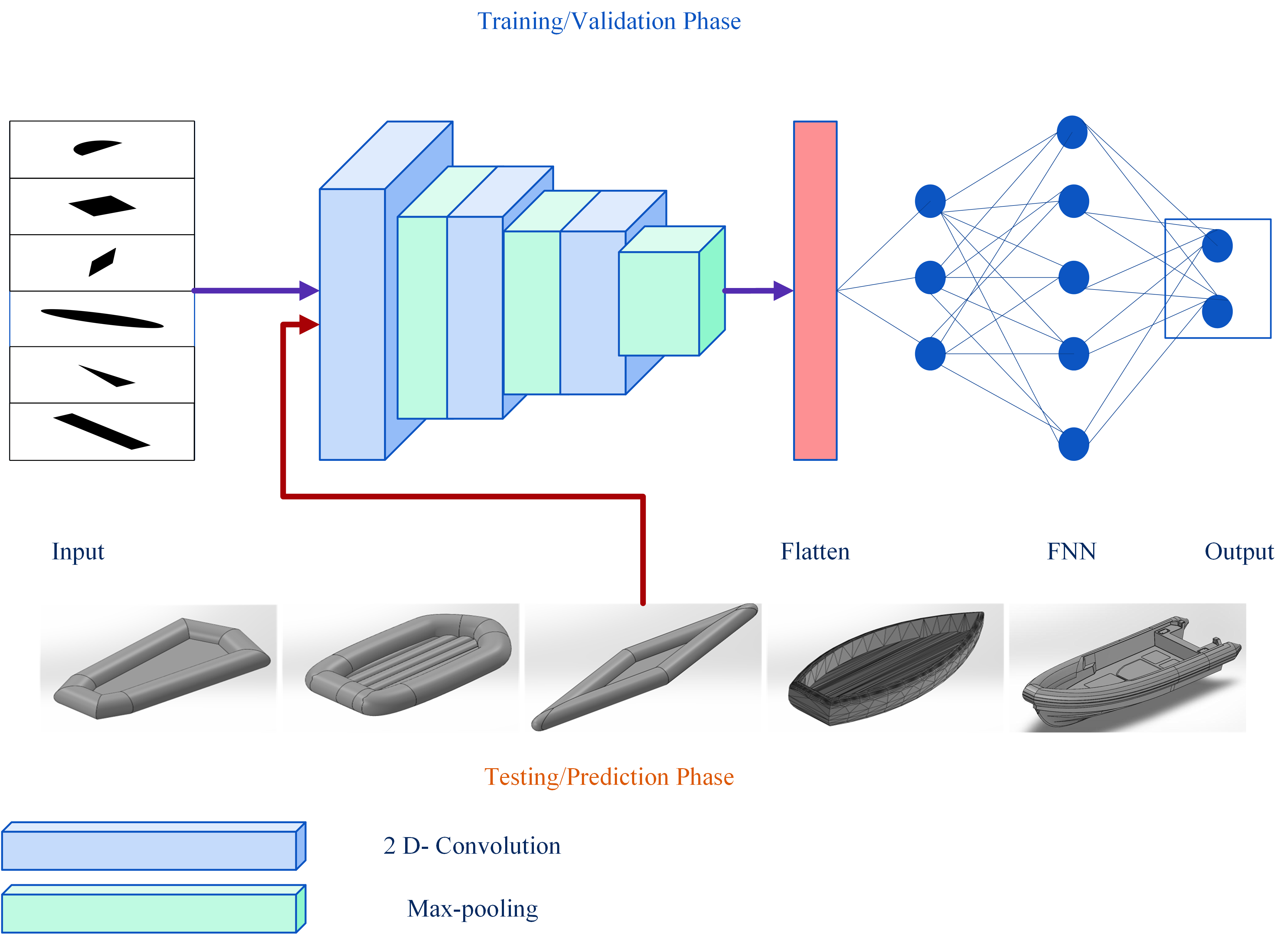}
     \vspace{-0.3in}
   \caption{Proposed convolutional neural network methodology for predicting the leeway objects' drag and lift coefficients} \label{fig:CNN}
\end{figure}

To efficiently estimate the coefficients governing the leeway objects utilized in this study, we propose a lightweight CNN architecture trained on labeled geometrical image data obtained in Subsection~\ref{sec:drag}. This trained model, illustrated in Figure~\ref{fig:CNN}, is used to predict the coefficients corresponding to the experimental objects. 

The train and validation datasets comprise input grayscale images of geometries,  resized to a fixed dimension of $128 \times 128$ pixels, along with their output drag ($C_D$) and lift ($C_L$) coefficients. The Conv2D layer implements a 2-D convolution operation, the MaxPool layer selects the maximum value within that region, and the Rectified Linear Unit activation function (ReLU) is defined as $f(x) = max(0,x)$. The output map is flattened into an input to the feedforward neural network. Denote the output of the third pooling layer as $X^c \;\in\;\mathbb{R}^{L_c\times W_c \times C_c}$, where $L_c$ is the height (length), $W_c$ is the width, and $C_c$ is the number of channels. We flatten by re-indexing all $L_cW_cC_c$ entries into a single column vector, denoted as 
\begin{align}
x^c \;=\;\mathrm{flatten}(X^c)\;\in\;\mathbb{R}^{L_cW_cC_c}.
\end{align}

The vector $x^c$ is then fed into two dense layers
\begin{align*}
 u^c  = \mathrm{ReLU}(W^c_1x^c +b^c_1)&\\
[\hat{C_D}, \hat{C_L}] = (W^c_2u^c +b^c_2)&,
\end{align*}

\noindent where weights $W^c_1, W^c_2$ and biases $b^c_1, b^c_2$ are learnable parameters.  Given $N_c$ numbers of input-output pair dataset $\{(I_i,\mathcal{O}_i)\}_{i=1}^{N_c}$, with targets $\mathcal{O}_i =[C_D, C_L]_i \in\mathbb{R}^2$, during training, we minimize the mean-squared error (MSE):
\begin{align}\label{eq:mse_loss}
\mathcal{L}_{\mathrm{MSE}} = \frac{1}{N} \sum_{i=1}^{N} \| \hat{\mathcal{O}}_i - \mathcal{O}_i \|^2_2.
\end{align}

The CNN model is optimized using the Adam~\cite{kingma2017adam} optimizer. Performance is evaluated using both MSE and Mean Absolute Error (MAE): 
\begin{align}\label{eq:mae_loss}
\mathcal{L}_\mathrm{MAE} = \frac{1}{N} \sum_{i=1}^{N} \bigl| \hat{\mathcal{O}}_i - \mathcal{O}_i \bigr|.
\end{align}

Once trained, the models are used to infer coefficients for our object images. Given images $I_\mathrm{test}$, the model outputs:
\begin{align}
f_{\theta}(I_\mathrm{test}) = [\hat{C}_D, \hat{C}_L]_\mathrm{test},
\end{align}

\noindent where $f_{\theta}$ represents the learned function parameterized by weights $\theta$ ( kernels, weights, and biases). The key layers of the CNN architecture are summarized in Algorithm~\ref{algm:cnn}. 

\begin{algorithm}[t]
\caption{CNN Architecture for Drag ($C_D$) and Lift ($C_L$) Coefficients Estimation}\label{algm:cnn}
\begin{algorithmic}[1]
\Inputs {Geometrical grayscale image $I\in\mathbb{R}^{H^c\times W^c\times1}$}
\Outputs {Predicted coefficients $[\hat{C}_D, \hat{C}_L]_\mathrm{test}$ \\ Trained weights $f_\theta$ }
\State $X \gets I$
\Initialize {$f_\theta$}
\While {not converged and training}
\State $X \gets \mathrm{Conv2D}(64,\;5\times5,\;\mathrm{ReLU})(X)$
\State $X \gets \mathrm{MaxPool2D}(2\times2)(X)$
\State $X \gets \mathrm{Conv2D}(32,\;3\times3,\;\mathrm{ReLU})(X)$
\State $X \gets \mathrm{MaxPool2D}(2\times2)(X)$
\State $X \gets \mathrm{Conv2D}(32,\;3\times3,\;\mathrm{ReLU})(X)$
\State $X^c \gets \mathrm{MaxPool2D}(2\times2)(X)$
\State $x^c \gets \mathrm{Flatten}(X^c)$
\State $u^c \gets \mathrm{Dense}(32,\;\mathrm{ReLU})(x^c)$
\State $[\hat C_D,\hat C_L] \gets \mathrm{Dense}(2)(u^c)$
\EndWhile
\State \Return $\hat C_D,\;\hat C_L,\;f_\theta$

\While {prediction}
\Inputs {$I_\mathrm{test}:$ Leeway object images }
\State $[\hat{C}_D, \hat{C}_L]_\mathrm{test} \gets f_{\theta}(I_\mathrm{test}) $
\EndWhile
\State \Return $[\hat{C}_D, \hat{C}_L]_\mathrm{test}$

\end{algorithmic}
\end{algorithm}

\subsection{Input Data Estimation Approach}
To enable accurate drift prediction, we first conducted controlled experiments to capture the key environmental and object-specific parameters influencing leeway motion such as the northward and eastward components of water current velocity $v_w = (v_{w,x}, v_{w,y})$, wind speed $v_a = (v_{a,x}, v_{a,y}))$, and the corresponding drift trajectories $\mathcal{Y} = (d^x,d^y)$. In addition, object parameters such as surface areas $A_a$ and masses ($m_o$) were measured. In addition to these experiments, we estimated drag, $C_D$, and lift, $C_L$, coefficients for each object. With these features, we consider some other calculated features, namely, air and water drag and lift forces $D_a = (D_{a,x}, D_{a,y})$, $D_w = (D_{w,x}, D_{w,y})$, $L_a = (L_{a,x}, L_{a,y})$, $L_w = (L_{w,x}, L_{w,y})$ and the rate of submersion, $\gamma$. 

\vspace{0.1in}
\subsubsection{Air and Water Drag Forces}
For many floating bodies, drift velocity arises from a balance between air and water drag forces~\cite{SOYLEMEZ1996423}. The drag force on the floating object due to air pulls the object in the wind direction, and the drag force on the floating object due to water resists motion through the water. The drag forces act in the opposite direction to fluid flow, and are modeled relative to the medium the object is moving through. The magnitude of the drag force equation is given by
\begin{align}\label{eq:drag:fa}
D_i (t) = -\frac{1}{2} \rho_i \cdot C_D^i \cdot A_i \cdot \tilde v_i^2 (t), \quad i \in \{a,w\}
\end{align}

\noindent where the notations $a$ and $w$ denotes air and water respectively, $\rho_a=1.225\,\mathrm{kg/m}^3$,  denotes the density of the air, $\rho_w=1025\,\mathrm{kg/m}^3$ is the density of water $C_D^i$ is the drag coefficient, $A_i$ is the surface area of the object facing the relative flow of the fluid and $\tilde v_i = v_i-v$ denotes the velocity of the object relative to the fluid $v_i$. The relative velocity vector components in the northward and eastward directions are $\tilde v_i(t) = (\tilde v_{i,x}(t), \tilde v_{i,y}(t))$ resulting in their magnitudes being estimated as {\small $\tilde v_i(t) = \sqrt{(\tilde v_{i,x}(t))^2 + (\tilde v_{i,y}(t))^2}$}.  Note that the eastward and northward components of the drag forces where $\cos(\theta) = \frac{\tilde{v}_{i,x}}{\tilde v_i}$ and $\sin(\theta) = \frac{\tilde{v}_{i,y}}{\tilde v_i}$ are $D_x = D\cos(\theta)$ and $D_y = D\sin(\theta)$. These result in the forces decomposition of Equation~\ref{eq:drag:fa} to 
\begin{align}\label{Eq:dragforceA}
D_{i,x}(t) &= -\frac{1}{2} \cdot \rho_i \cdot C_D^i \cdot A_i \cdot \tilde v_i (t) \cdot \tilde v_{i,x} (t) \nonumber\\
D_{i,y}(t) &= -\frac{1}{2} \cdot \rho_i \cdot C_D^i \cdot A_i \cdot \tilde v_i (t) \cdot \tilde v_{i,y} (t) \quad i \in \{a,w\}.
\end{align}

\vspace{0.1in}

\subsubsection{Air and Water Lift Force}
In contrast to drag, which acts in the direction opposed to motion, lift acts perpendicular to the motion. Lift force follows a similar and prependicular estimation as the drag forces where instead $\tilde v_{i,\perp} = \left(-\frac{\tilde{v}_{i,y}}{\tilde v_i} \frac{\tilde{v}_{i,x}}{\tilde v_i}\right)$  and the decomposed forces result in 
\begin{align}\label{eq:lift:fa}
L_i (t) &= \frac{1}{2} \rho_i \cdot C_L^i \cdot A_i \cdot \tilde v_i^2 (t) \tilde v_{i,\perp} (t) 
\nonumber\\
&= \frac{1}{2} \rho_i \cdot C_L^i \cdot A_i \cdot \tilde v_i (t)\left(-\tilde{v}_{i,y}, \tilde{v}_{i,x}\right)\quad i \in \{a,w\}
 \end{align}
 
and are decomposed into $x,y$ components
\begin{align}
L_{i,x}(t) &= -\frac{1}{2} \cdot \rho_i \cdot C_L^i \cdot A_i \cdot \tilde v_i (t) \cdot \tilde v_{i,y} (t) \nonumber\\
L_{i,y}(t) &= \frac{1}{2} \cdot \rho_i \cdot C_L^i \cdot A_i \cdot \tilde v_i (t) \cdot \tilde v_{i,x} (t) \quad i \in \{a,w\}.
\end{align}

\vspace{0.1in}

\subsubsection{Leeway Objects' Submersion Rate}
The extent to which an object is submerged in water plays an important role in determining both its buoyancy and hydrodynamic resistance, thus directly influencing its drift behavior. The submersion rate controls the balance between water-following and wind-following behavior and is simply defined as 
\begin{align}
\gamma = \frac{A_w}{A_w + A_{a}}.
\end{align}

In summary, numerically, we introduced calculated features in addition to the experimental and physical ones are summarized in Table~\ref{tab:vars}, reporting their symbols, descriptions, number of dimensions (Dim.), and units. In Appendix~\ref{App:A}, we plot the numerical input data with respect to time.

\begin{table}[!ht]
\caption{Summary of the numerical input features}\label{tab:vars}
\centering
\resizebox{1\linewidth}{!}{
\begin{tabular}{|p{1.8cm}|c|l|c|c|}
\hline
Input Type&  Symbol & Description & Dim.  & Units   \\ \hline \hline
\multirow{7}{2cm}{Time-varying input variables} &$v_a$& Wind velocity at $10\,\mathrm{m}$ height & $2$& $\mathrm{m/s}$\\
& $v_w$     & Water current velocity& $2$&$\mathrm{m/s}$ \\                                                  
& $D_{a}$& Drag force from wind on object &$2$& $N$     \\
& $D_{w}$      & Drag force from water on object&$2$& $N$    \\ 
& $L_{a}$& Lift force from wind on object &$2$& $N$     \\
& $L_{w}$      & Lift force from water on object& $2$&$N$  \\ 
&$T$& Time&1&$s$\\ \hline
\multirow{2}{2cm}{Object-specific constant inputs}& 
$m_o$  & Mass of the drifting object   &$1$ & $kg$    \\
& $\gamma$    & Submersion rate &$1$& -  \\
\hline
Target outputs& $\mathcal{Y}$  & Drift &$2$   & $\mathrm{m}$    \\
\hline
\end{tabular}}
\end{table}

\vspace{-0.1in}
 \section{Multi-Modal Drift Forecasting with Attention-Based Sequence-to-Sequence Models}
\begin{figure*}[t]
    \centering
    \includegraphics[width =\linewidth]{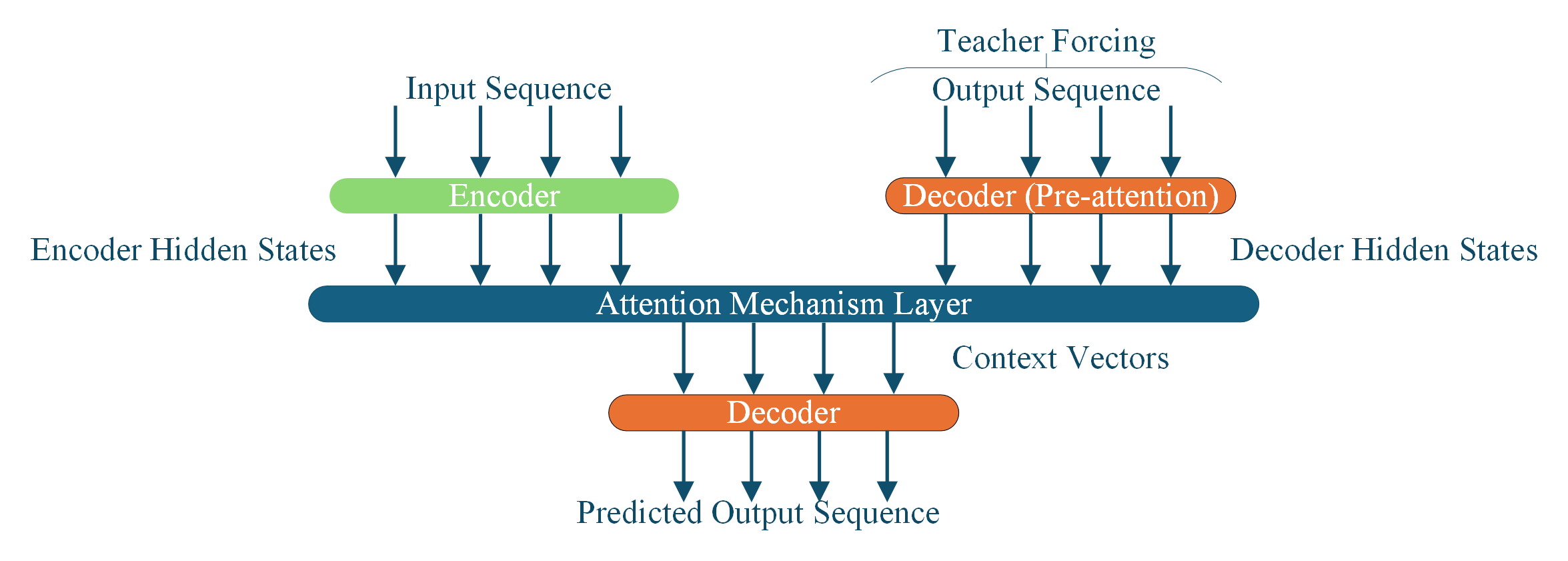}
     \vspace{-0.3in}
   \caption{Proposed Attention-Based Sequence to Sequence Method for predicting the leeway drift} \label{fig:seqeq}
\end{figure*}

In this section, we formulate the multi-modal drift prediction framework that integrates an LM in the form of a Sentence Transformer with attention-based sequence-to-sequence prediction models. We present a three-step framework to forecast the trajectory sequence of a drifting object, $\mathcal{Y}_{t+1: t+\ell_d}$, from past inputs $\mathcal{X}_{t-\ell_e+1:t}$: (i) embedding object-specific textual descriptions with a Sentence Transformer and fusing these embeddings with environmental and physical time series; (ii) attention-based sequence-to-sequence LSTM (MM-Attention-STS-LSTM); and (iii) an alternative sequence-to-sequence Transformer (MM-STS-Transformer) to assess robustness. The general concept of both models is illustrated in Figure~\ref{fig:seqeq}.

\subsection{Multi-Modal Input Data using Language Models}
To incorporate qualitative object information into our multi-modal drift-prediction framework, we propose a multi-modal input data consisting of time-series observations and object-level semantic descriptors. We first assembled concise natural-language descriptions for each leeway object, summarizing key geometric and material attributes (e.g., "Inflatable orange raft with rounded front and flat rear. Constructed of lightweight PVC with no canopy."). These descriptions capture salient visual and structural characteristics that are not easily expressed through numerical sensor readings alone. We then use a pre-trained, sentence-transformer model~\cite{reimers-2019-sentence-bert} (\texttt{all-MiniLM-L6-v2}) to map each textual description into a fixed-length, $384$-D embedding vector. This model, based on a Siamese network architecture fine-tuned on large paraphrase and semantic-similarity datasets, produces embeddings in which semantically similar descriptions lie close in Euclidean space. Concretely, for each description $d_i$, we compute
\begin{align}\label{eq:st}
\mathbf{e}_i \;=\; \mathrm{ST}(\,d_i\,)\;\in\;\mathbb{R}^{384},
\end{align}

\noindent where $\mathrm{ST}(\cdot)$ is the Sentence Transformer's forward pass and $\mathbf{e}_i$ denotes object $i$'s textual metadata. Finally, we concatenated these $384$-D vectors with the corresponding time series for each object, yielding a multi-modal input $\mathcal{X}\in \mathbb{R}^{N\times 399}$ for our model. By embedding the textual modality in this manner, the network can learn cross-modal correlations, improving its ability to generalize across heterogeneous object types. 
Recall that the original time length of data, $T_\mathrm{original}$, is down-sampled at $t_h$ time horizons to yield a time length $T$. The multi-model input data preprocessing is described in Algorithm~\ref{algm:2}. 

\begin{algorithm}[ht]
\caption{Multi-Modal Input Data Processing}
\label{algm:2}
\begin{algorithmic}[1]
\Inputs{Objects' textual description $\{d_i\}_{i=1}^5$, numerical input data $TD$, output $\mathcal{Y}$, encoder length $\ell_e$, decoder forecast length $\ell_d$, time horizon $t_h$}
\Outputs{Sequence dataset $ \{\mathcal{X}^e\},\;\{\mathcal{Y}^{d}\},\;\{\mathcal{Y}^{\mathrm{out}}\}$}
  \For{$i=1$ \textbf{to} $5$} \Comment{Number of objects}
    \State $\mathbf{e}_i \gets$ Eq.~\ref{eq:st}   
    \State $\{\mathcal{X}\}_i \gets [\,TD_i,\;\mathbf{e}_i\,]\in\mathbb{R}^{N\times 399}$ \Comment{$p=399$ for brevity}
    \State $\{\mathcal{X}\}_i \gets \{\mathcal{X}\}_i[::t_h]\in\mathbb{R}^{N//t_h\times p}$ \Comment{Sample at $t_h$}
    \State Input 
    $\{\mathcal{X}^e_i\} \gets$ Eq.~\ref{eq:inseq} $\!\in\!\mathbb{R}^{(N//t_h - \ell_e ) \times \ell_e\times p}$ 
    
  \State $\{\mathcal{Y}\}_i \gets \{\mathcal{Y}\}_i[::t_h]\in\mathbb{R}^{N//t_h\times 2}$ 

     \State $\{\mathcal{Y}^\mathrm{out}_i\} \,, \{\mathcal{Y}^d_i\} \gets$ Eqs.~\ref{eq:outseq} and \ref{eq:tf} $\!\in\!\mathbb{R}^{(N//t_h -\ell_d)\times \ell_d \times 2}$
  \EndFor
  
  \State Concatenate  $\{\mathcal{X}^e_i\}$, $\{\mathcal{Y}^d_i\}, \{\mathcal{Y}^\mathrm{out}_i\}$ time-wise and store $\{\mathcal{X}^e\},\;\{\mathcal{Y}^d\},\;\{\mathcal{Y}^{\mathrm{out}}\}$ 
\end{algorithmic}
\end{algorithm}

\vspace{-0.1in}
\subsection{Attention-based Sequence-to-sequence LSTM}
To train our model to map past observations onto future drift trajectories, we first employ sequence-to-sequence autoencoder models with teacher forcing. Let $\mathcal{X}\in \mathbb{R}^{T\times p}$ be the fmulti-modal time-series, $\mathcal{Y}\in\mathbb{R}^{T\times 2}$ be the corresponding drift ground-truth time series, $\ell_e$ the encoder window length, and $\ell_d$ the prediction horizon. For $N - (\ell_e +\ell_d ) + 1$, the number of valid training examples, we can extract the encoder input sequence $\mathcal{X}^e_t$ at training example $t$
 \begin{align}\label{eq:inseq}   \mathcal{X}^e_i = \mathcal{X}\bigl[i : i+\ell_e\;\bigr]\;\in\mathbb{R}^{\ell_e\times p},
   \end{align}

\noindent where the decoder target
   \begin{align}\label{eq:outseq}
     \mathcal{Y}^{\mathrm{out}}_i = \mathcal{Y}\bigl[i+\ell_e : i+\ell_e+\ell_d\;\bigr]\;\in\mathbb{R}^{\ell_d\times 2},
   \end{align}
with the decoder teacher forcing input
   \begin{align}\label{eq:tf}
     \mathcal{Y}^d_i = 
     \begin{bmatrix}
       \mathbf{0}\\
       \mathcal{Y}[i+\ell_e : i+\ell_e+\ell_d-1,\;]
     \end{bmatrix}
     \;\in\mathbb{R}^{\ell_d\times 2},
   \end{align}

\noindent where the first row is a zero "start token" and the remaining rows are the true future drift offset by one timestep. These triplets are the inputs and output for our training. The first model utilizes LSTM layers. In general, at time $t$, LSTM takes an input vector $a_t$ and the previous hidden and cell states $(h_{t-1},c_{t-1})$, and computes:
$$
\begin{aligned}
i_t &= \sigma\bigl(W_i\,a_t \;+\; U_i\,h_{t-1} \;+\; b_i\bigr)   &&\text{(input gate)}\\
f_t &= \sigma\bigl(W_f\,a_t \;+\; U_f\,h_{t-1} \;+\; b_f\bigr)   &&\text{(forget gate)}\\
o_t &= \sigma\bigl(W_o\,a_t \;+\; U_o\,h_{t-1} \;+\; b_o\bigr)   &&\text{(output gate)}\\
\tilde c_t &= \tanh\bigl(W_c\,a_t \;+\; U_c\,h_{t-1} \;+\; b_c\bigr) &&\text{(cell candidate)}\\
c_t &= f_t \odot c_{t-1}\;+\;i_t \odot \tilde c_t                      &&\text{(cell state update)}\\
h_t &= o_t \odot \tanh(c_t)                                        &&\text{(hidden state)}
\end{aligned}
$$

Here, $\sigma(\cdot)$ is the sigmoid function, $\odot$ denotes element-wise multiplication, $W_{i,f,o,c}, U_{i,f,o,c}$, and $b_{i,f,o,c}$ are learnable weights and biases. The gating mechanisms allow the LSTM to learn long-term dependencies. For time $t=1,\dots,\ell_e$ of input vector $\mathcal{X}^e_t$, the 2-layer LSTM encoder is calculated as follows:
\begin{align}\label{eq:enc}
(h^{(1)}_t,\;c^{(1)}_t)&=\mathrm{LSTM}^{(1)}\bigl(\mathcal{X}^e_t,\;h^{(1)}_{t-1},\;c^{(1)}_{t-1}\bigr)\nonumber\\
(h^{(2)}_t,\;c^{(2)}_t)&=\mathrm{LSTM}^{(2)}\bigl(h^{(1)}_t,\;h^{(2)}_{t-1},\;c^{(2)}_{t-1}\bigr).
 \end{align}
 
The second layer retain its final hidden and cell states $\bigl(h^{(2)}_{\ell_e},\,c^{(2)}_{\ell_e}\bigr)$ to initialize the decoder, which comprises another LSTM layer, unrolled for $\ell_d$ steps in the future, and driven by the teacher-forcing inputs $\mathcal{Y}^d$. The decoder LSTM is run for $u=1,\dots,\ell_d$ steps with teacher forcing:
   \begin{align}\label{eq:dec}
     (h^{(3)}_u,\,c^{(3)}_u)
     &= \mathrm{LSTM}^{(3)}\bigl(\mathcal{Y}^d_u,\;d_{u-1},\,c'_{u-1}\bigr),
     \\
\bigl(h^{(3)}_{0},c^{(3)}_0\bigr)&=\bigl(h^{(2)}_{\ell_e},c^{(2)}_{\ell_e}\bigr).\nonumber
 \end{align}

Each decoder hidden state $h^{(3)}_u\in\mathbb{R}^{64}$ is then projected to the 2-D drift prediction via a shared linear map.

In the attention variant, the decoder is modified to attend over the encoder's hidden-state sequence at every timestep. This allows each decoded step to weigh all past encoder states adaptively, rather than relying solely on the fixed-size final state $(h_{\ell_e},c_{\ell_e})$. For each decoder step $u$, we define key,value and query for weights  $W^Q,W^K,W^V\in\mathbb{R}^{64\times d_k},\;d_k=32$:
   \begin{align}\label{eq:kqv}
     q_u = h^{(3)}_u\,W^Q,\quad k_t = h^{(2)}_t\,W^K,\quad      v_t = h^{(2)}_t\,W^V, 
 \end{align}

 apply a scaled dot-product attention
   \begin{align}\label{eq:sdatt}
     a_{u,t}
     = \frac{\exp\bigl(q_u\cdot k_t^\top/\sqrt{d_k}\bigr)}
            {\sum_{t'=1}^{\ell_e} \exp\bigl(q_u\cdot k_{t'}^\top/\sqrt{d_k}\bigr)},
     \quad
     c_u = \sum_{t=1}^{\ell_e} a_{u,t}\,v_t, 
   \end{align} 

and concatenate with the decoder state,
   $\bigl[h^{(3)}_u\;\|\;c_u\bigr]\in\mathbb{R}^{64+32}$, and pass through the same dense output:
    \begin{align}\label{eq:lstm:out}
     \hat y_{u} 
     &= W^\mathrm{out}\,\bigl[h^{(3)}_u\|c_u\bigr] +b^\mathrm{out},
     \quad 
     W^\mathrm{out} \in\mathbb{R}^{2\times 96},\nonumber
      \\
     \hat{\mathcal{Y}}^{\mathrm{out}} &=[\,\hat y_1,\dots,\hat y_{\ell_d}\,].
   \end{align}

This attention-based sequence-to-sequence LSTM trained with the multi-model input data forms our multi-modal model named MM-Attention-STS-LSTM and is summarized in Algorithm~\ref{algm:3}.

\begin{algorithm}[!ht]
\caption{Multi-Modal Attention-based Sequence-to-Sequence LSTM}\label{algm:3}
\begin{algorithmic}[1]
\Inputs{Multi-modal training input: $\mathcal X^{e}$, decoder input: $\mathcal Y^d$} 
\Initialize{$(h_0,c_0) \gets (0,0)$, all weights}
\Outputs{Predicted sequence: $\hat {\mathcal{Y}}^{\mathrm{out}}$ \\ Trained weights: $f_\theta$ }
\While {not converged and training}
\State \textbf{--- Encoder ---}
\For{$t=1$ to $\ell_e$}
\State $(h^{(2)}_t,c^{(2)}_t) \gets \mathcal X^{e} $ Eq.~\ref{eq:enc}  
\EndFor
\State Store encoder outputs $\mathcal{H} = [\,h^{(2)}_1,\dots,h^{(2)}_{\ell_e}\,]$
\State \textbf{--- Decoder ---}
\Initialize {Decoder state $(h^{(3)}_0,c^{(3)}_0) \gets (h^{(2)}_{\ell_e},c^{(2)}_{\ell_e})$}
\For{$u=1$ to $\ell_d$}
  \State $(h^{(3)}_u,c^{(3)}_u)\;\gets\;$ Eq.~\ref{eq:dec}
  \State Projections
    $q_u, \, k_t, \, v_t\, \gets$ Eq.~\ref{eq:kqv} 
  \State Attention and context vector $c_u \gets$  Eq.~\ref{eq:sdatt} 
  \State $\hat y_u \gets$ Eq.~\ref{eq:lstm:out}
\EndFor
\EndWhile
\State \Return $f_{\theta}$

\While {prediction}
\Inputs {Multi-modal test input: $\mathcal X^{e}$}
\State $\hat{\mathcal{Y}}_\mathrm{out} \gets f_{\theta}(\mathcal X^{e})$ 
\EndWhile
\State \Return $\hat{\mathcal{Y}}_\mathrm{out}$
\end{algorithmic}
\end{algorithm}

 \subsection{Multi-Modal Drift Forecasting with Sequence-to-Sequence Transformer}

In addition to MM-Attention-STS-LSTM, we develop a transformer-based sequence-to-sequence model that maps the multi-modal input sequence $\mathcal{X}^e$ to an output sequence $\mathcal{Y}^{\mathrm{out}}$. The model consists of an encoder and a decoder, each comprising multi-head self-attention, causal mask, feedforward layers, and layer normalization.
First, the encoder projects the input sequence $\mathcal{X}^e$ into a $d_m = 64$ dimensional space using a linear embedding layer. A fixed sinusoidal positional encoding $\mathrm{PE}(t,k)$ is then added to the embedded input, where  
\begin{align}
\mathrm{PE}(t, 2i)   &= \sin\left(\frac{t}{10000^{\frac{2i}{d_m}}}\right), \nonumber\\
\mathrm{PE}(t, 2i+1) &= \cos\left(\frac{t}{10000^{\frac{2i}{d_m}}}\right),
\end{align}
injecting temporal information into each embedding dimension. 
Layer normalization is then applied to the sum of the embedding and its positional encoding, producing $z^{(0)}_t$, which serves as the input to the encoder:
\begin{align}\label{eq:project}
z^{(0)}_{t} &= \mathrm{LayerNorm}(W^E\,\mathcal{X}^e_t + b^E + \mathrm{PE}(t), \nonumber\\
Z^{(0)} &= [\,z^{(0)}_{1},\dots,z^{(0)}_{\ell_e}\,].
\end{align}

Given $Z^{(0)}\in\mathbb{R}^{\ell_e\times d_m}$, the encoder performs a multi-head attention by computing key, value and query projections as described in Equation ~\ref{eq:kqv}, where $Z^{(0)}$ replaces $h^{(2)}$ and $h^{(3)}$. For each head $h = 1,\dots, H$, we compute the scaled dot attention, $\mathrm{head}_h$ as described in Equation~\ref{eq:sdatt}. These outputs are then concatenated and projected using a learnable weight $W^H\in\mathbb R^{H\,d_k\times d_m}$:
\begin{align}\label{enc:MHA}
\mathrm{MHA}(Z^{(0)}) = [\mathrm{head}_1;\dots;\mathrm{head}_H]\,W^H.
\end{align}
The attention layer is succeeded by a layer normalization $U^e = \mathrm{LayerNorm}(\mathrm{MHA}(Z^{(0)})$, which is then fed into a position-wise feedforward neural network and its layer normalization, yielding encoder outputs $Z^{(1)}$ : 
\begin{align}\label{eq:encffn}
\mathrm{FFN}_e &= W^E_2\,\mathrm{ReLU}(W^E_3\,U^e + b^E_3) + b^E_2, \nonumber\\
Z^{(1)} &= \mathrm{LayerNorm}\bigl(\mathrm{FFN}_e\bigr), \; \in \mathbb{R}^{\ell_e\times d_m}.
\end{align}
with $W^E_3\in\mathbb R^{d_m\times f},\,W^E_2\in\mathbb R^{f\times d_m}$.

Note that $f$ is the feedforward layer dimension. To teacher force, the decoder takes as input the (shifted) target sequence $\mathcal{Y}^d$ as described in Equation~\ref{eq:inseq}. The decoder also starts by embedding and adding a positional encoding to $\mathcal{Y}^d$ to output $D^{(0)}\in\mathbb R^{\ell_d\times d}$. The reason is that, just like the encoder, the decoder has no built-in sense of time order, so we must inject position information. We perform a masked self-attention (MSA) operation on $D^{(0)}$ by first computing key values and query projections of $D^{(0)}$, and further explicitly defining  
\begin{align}\label{eq:MSA}
\alpha_{ij}
= \frac
   {\exp\!\Bigl(\tfrac{q_i\cdot k_j}{\sqrt{d_k}}\Bigr)\,\mathbf{1}_{j\le i}}
   {\displaystyle\sum_{j'=1}^{T}\exp\!\Bigl(\tfrac{q_i\cdot k_{j'}}{\sqrt{d_k}}\Bigr)\,\mathbf{1}_{j'\le i}}
,\;
\mathrm{head}(D^{(0)})_i =\sum_{j=1}^{T}\alpha_{ij}\,v_j,\nonumber\\
\mathrm{MHA}(D^{(0)}) = [\mathrm{head}(D^{(0)})_i;\dots;\mathrm{head}(D^{(0)})_H]\,W^H_d.
    \end{align}

\noindent where $\mathbf{1}_{j\le i}$ enforces the causal mask by only attending to positions $\le i$). The multihead masked attention layer is followed by an add and layer normalization operation to yield $U^d = \mathrm{LayerNorm}(D^{(0)} +  \mathrm{MHA}(D^{(0)})$. The decoder leverages information of the encoder outputs $Z^{(1)}$ by performing multihead cross attention mechanism operation:
    \begin{align}\label{eq:crossatt}
      Q^d_2 = U_d \,W^{Q^d_2},\;
      K^d_2 = Z^{(1)}\,W^{K^d_2},\;
      V^d_2 = Z^{(1)} \,W^{V^d_2},
  \nonumber\\
      A^d_2 = \mathrm{softmax}\!\bigl(\frac{Q^d_2(K^d_2)^\top}{\sqrt{d_k}}\bigr)\,V^d_2.
    \end{align}
Add layer $A^d_2+ U_d$, which is normalized to yield $U_d^2 \in\mathbb{R}^{\ell_d\times d_m}$, which is finally mapped back to output dimension space with the following mapping :
\begin{align}
\mathrm{FFN}_d = W^D_2\,\mathrm{ReLU}(W^D_1\,U_d^2 + b^D_1) + b^D_2,\\
D^{(1)} = \mathrm{LayerNorm}\bigl(U_d^2 + \mathrm{FFN}_d\bigr),  \label{ffn_d}\\
\hat{\mathcal{Y}}^{\mathrm{out}} = W^P\,D^{(1)} + b^P.\label{eq:trout}
\end{align}

This model is summarized in Algorithm~\ref{algm:4}.

\begin{algorithm}[!ht]
\caption{Multi-Modal Drift Forecasting with Sequence-to-Sequence Transformer}\label{algm:4}
\begin{algorithmic}[1]
\Inputs{Multi-modal training input: $\mathcal X^{e}$, decoder input: $\mathcal Y^d$} 
\Outputs{Predicted sequence $\hat {\mathcal{Y}}^{\mathrm{out}}$\\ Trained weights: $f_\theta$}
\Initialize {all weights}
\While {not converged and training}

\State \textbf{--- Encoder ---}
\State $Z^{(0)}\gets$ Eq.~\ref{eq:project}\Comment{Input Embedding + Positional Encoding + Layer norm} 
\State $Z^{(0)} \gets \mathrm{LayerNorm}(Z^{(0)})$

\State $\mathrm{MHA}(Z^{(0)})\gets$ Eqs.~\ref{eq:kqv},~\ref{eq:sdatt},~\ref{enc:MHA} \Comment{Multihead  Self-Attention}
    \State $U^e \gets \mathrm{LayerNorm}(\mathrm{MHA}(Z^{(0)})$
  \State $Z^{(1)}\gets$ Eq.~\ref{eq:encffn} 
\State \textbf{--- Decoder ---}
 \State $D^{(0)}\gets$ Eq.~\ref{eq:project}\Comment{Output Embedding + Positional Encoding + Layer norm} 

 \State $\mathrm{MHA}(D^{(0)})\gets$  Eq.~\ref{eq:MSA} \Comment{Multihead masked Self-Attention}
\State $U^d \gets \mathrm{LayerNorm}(D^{(0)} +  \mathrm{MHA}(D^{(0)})$ \Comment{Add and layer norm}
 \State $A^d_2 \gets$  Eq.~\ref{eq:crossatt}
\State $U_d^2 \gets \mathrm{LayerNorm}(A^d_2+ U_d)$ \Comment{Add and layer norm}
\State $D^{(1)} \gets$ Eq.~\ref{ffn_d}
\State $\hat{\mathcal{Y}}_\mathrm{out} \gets$ Eq.~\ref{eq:trout}
\EndWhile
\State \Return $f_{\theta}$

\While {prediction}
\Inputs {Multi-modal test input: $\mathcal X^{e}$}
\State $\hat{\mathcal{Y}}_\mathrm{out} \gets f_{\theta}(\mathcal X^{e})$ 
\EndWhile
\State \Return $\hat{\mathcal{Y}}_\mathrm{out}$
\end{algorithmic}
\end{algorithm}

At test time, we remove teacher forcing and generate the output auto-regressively. In training, we minimize the MSE between $\hat{\mathcal{Y}^\mathrm{out}}$ and the true target sequence $\mathcal{Y}^\mathrm{out}$ as described in Equation~\ref{eq:mse_loss}. Optimization is performed with Adam and learning rate $\xi$. After training, we evaluate each model on the held-out validation split by computing Root Mean Squared Error (RMSE) $= \sqrt{\mathrm{MSE} }$, MAE as described in Equation~\ref{eq:mae_loss}, and the Mean Absolute Percentage Error (MAPE)
  \begin{align} \frac{100\%}{N}\sum_{i,t}\!\frac{\bigl|\hat{ \mathcal{Y}}^{\mathrm{out}}_{i,t} - \mathcal{Y}^{\mathrm{out}}_{i,t}\bigr|}{|\mathcal{Y}^{\mathrm{out}}_{i,t}|}.  \end{align}
\section{Results of Proposed Methods}\label{Sec:Results}
We discuss results in two main parts: (i) objects' drag and lift coefficients estimation and (ii) multi-modal drift prediction results. All numerical experiments discussed in this section were performed on a single workstation running \texttt{Windows~11}, equipped with a 2.21\, GHz Intel(R) Xeon(R) W9-3475X, 256\, GB RAM, and an NVIDIA RTX 4500 Ada Generation (24\, GB WDDM). The core modeling and training routines were implemented in \texttt{Python~3.9.21}, using \texttt{TensorFlow 2.10.1} with \texttt{CUDA 11.2} and \texttt{cuDNN 8.0} for GPU acceleration.

\subsection{Drag and Lift Coefficients Estimation Results}
We trained the proposed lightweight CNN on a total of $179$ geometrical image datasets with a learning rate of $0.001$ and a train-validation split of $90\% - 10\%$ for performance monitoring. To prepare the data, images are normalized to the $[0, 1]$ range using the following equation
\begin{align}
     I = \frac{I{\mathrm{original}}}{255} .
\end{align}

Training is conducted over $200$ epochs with a batch size of $32$ and an early stopping criterion (patience =$30$) on the validation MSE loss. The trained model is then used to predict the $C_D$  and $C_L$ values of the leeway object images given the CAD models as input. We compare a lightweight CNN and a CNN that combines attention and positional encoding layers. To build the attention-based CNN, instead of flattening $X^c$, it is reshaped into an input sequence for the attention as 
$X_{\mathrm{flat}} = \mathrm{Reshape}(X^c) \in \mathbb{R}^{S^c \times C^c}$, where $S^c  = L^c \times W^c$. We add a learned positional embedding of shape $S^c \times C^c$ so the attention layer is aware of spatial order. The positional encoding is initialized with a random normal distribution and is learnable. We then add a learned positional embedding matrix to $X_{\mathrm{flat}}$ and proceed to compute the multi-head attention for the per-head key/query dimension $d_k =32$. The output of the multi-head attention is then aggregated via global average pooling. Fully connected layers map the aggregation to $C_D$ and $C_L$. We trained models with $1$, $2$, $4$, and $8$ heads in addition to the proposed version. Results are reported in Table~\ref{tab:clcdval}.  

\begin{table}[h]
    \caption{Comparison of Validation‐set Performance of the Lightweight CNN and the Attention‐Augmented CNNs}
    \label{tab:clcdval}
   \centering
    \begin{tabular}{|c|c|c|}
    \hline
    Model	& MSE &	MAE \\ \hline \hline
    Lightweight CNN &$1.5537$ &$ 0.9239$ \\ \hline
    CNN + Attention ($1$ head) &  $8.9316$ & $2.4267$ \\ \hline
        CNN + Attention ($2$ heads) &$2.3282$& $1.1990$\\ \hline
        CNN + Attention ($4$ heads) &$6.0433$& $1.8047$ \\ \hline
        CNN + Attention ($8$ heads) & $5.0413$& $1.5854$\\ \hline
    \end{tabular}
\end{table}

\begin{figure}[ht]
    \centering \includegraphics[width=1\linewidth]{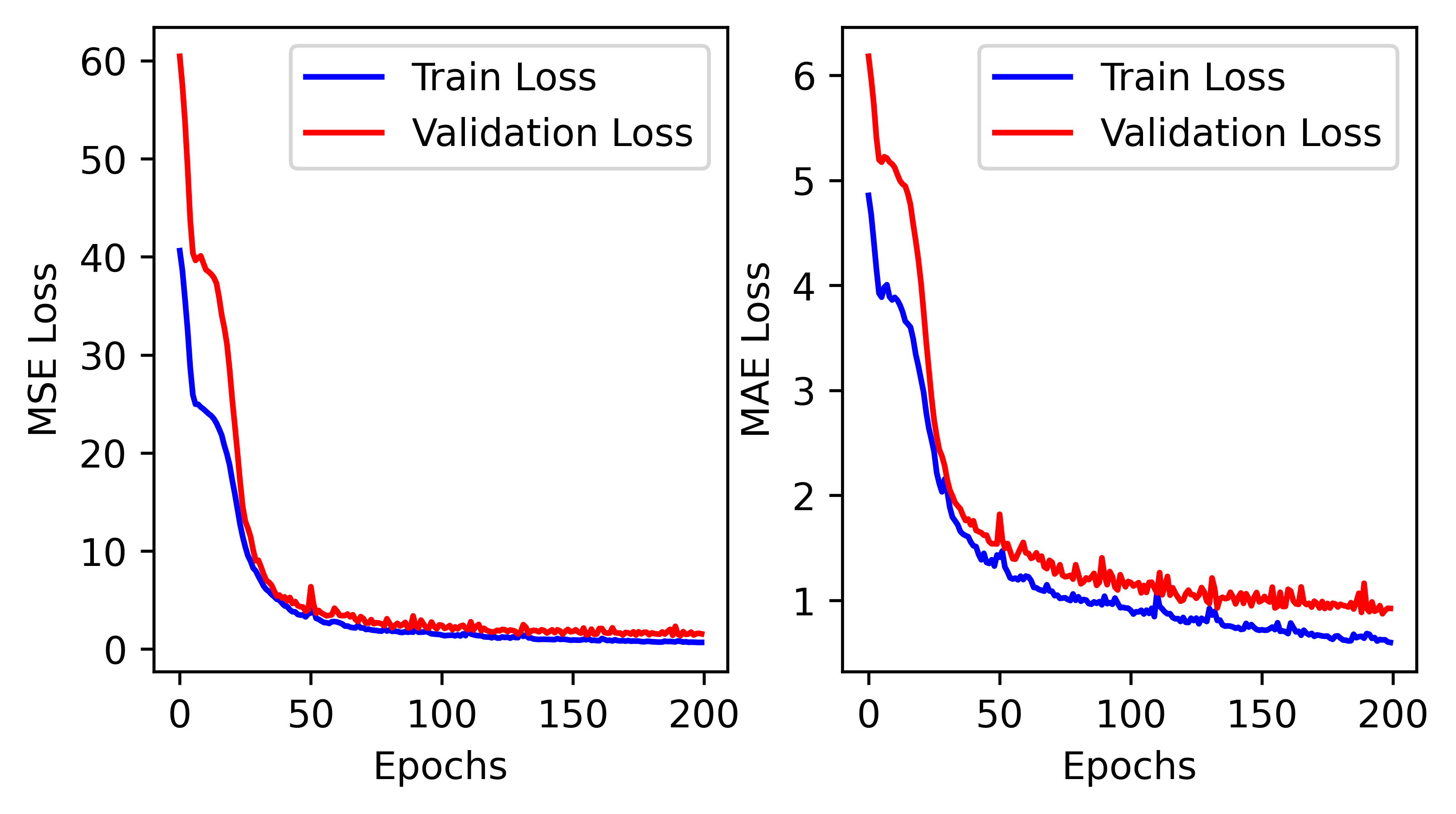}
     \vspace{-0.3in}
    \caption{Training and Validation MSE and MAE Losses of Lightweight CNN}\label{fig:CNN_Loss}
\end{figure}

On the test set, two models are most notable as they achieved the lowest scores: the model with a $2$- head attention achieved an MSE loss of $2.3282$ and an MAE $1.1990$. The proposed lightweight CNN achieves an MSE of $1.5537$ and MAE of $0.9239$, outperforming $2$-head-attention‐augmented CNN by $\approx0.8$ in MSE and $\approx 0.2$ in  MAE. The superior performance of the lightweight CNN may be due to the relatively small dataset size( $179$). The lightweight architecture has fewer parameters, reducing the risk of overfitting. In contrast, the attention-augmented CNN introduces additional complexity that may not be fully exploited with the limited available data, resulting in weaker generalization. Figure~\ref{fig:CNN_Loss} illustrates the convergence of train versus validation losses of the lightweight CNN, with both losses decreasing steadily, highlighting stable optimization and minimal overfitting.

\begin{figure}[!ht]
    \centering
    \includegraphics[width=1\linewidth]{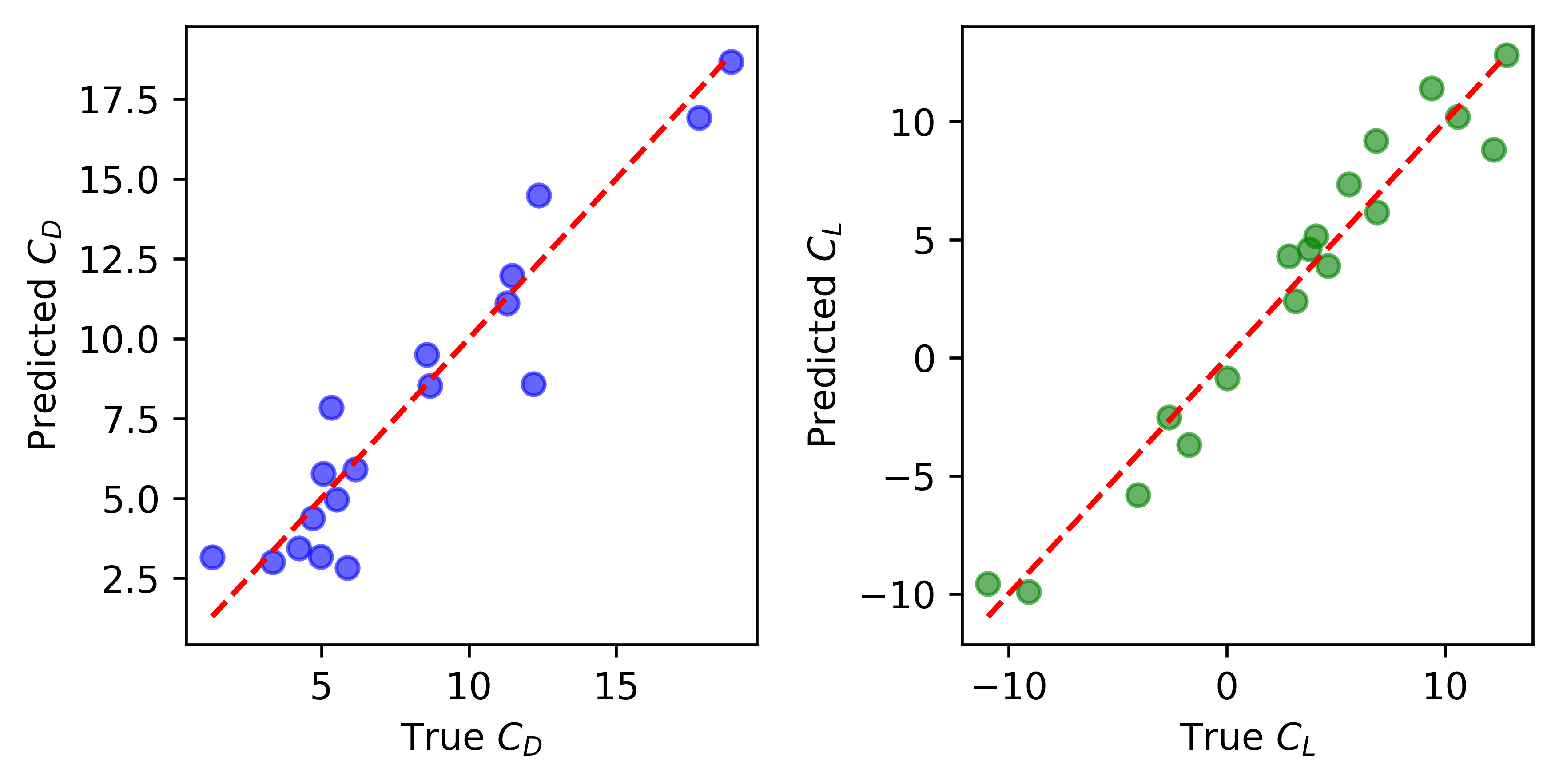}
     \vspace{-0.3in}
    \caption{True vs Predicted Validation $C_D$ and $C_L$ with $2$-head-attention-based CNN}
    \label{fig:Val-AttCNN}
\end{figure}

\begin{figure}[!ht]
    \centering
    \includegraphics[width=1\linewidth]{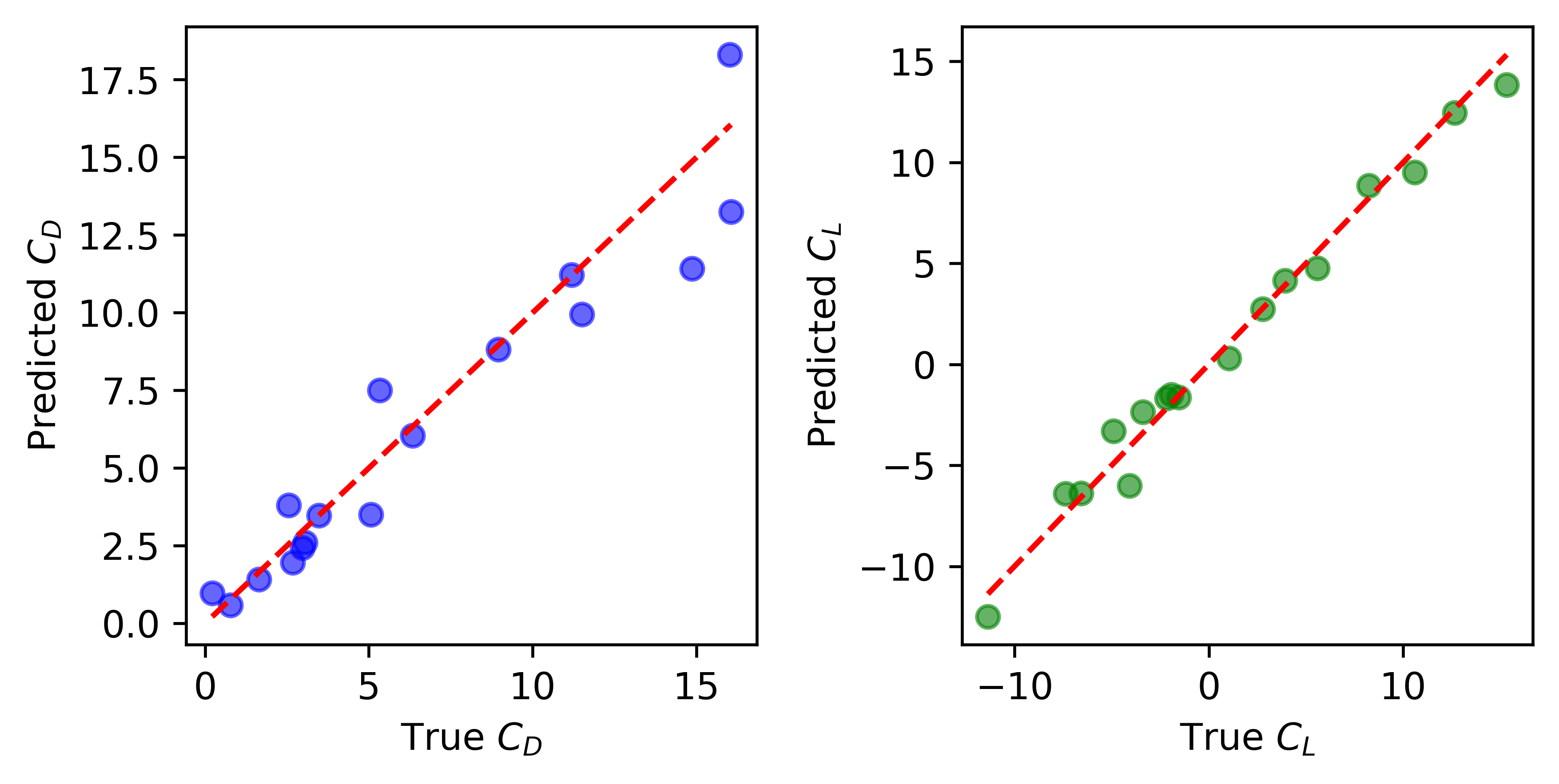}
     \vspace{-0.3in}
    \caption{True vs Predicted Validation $C_D$ and $C_L$ with lightweight CNN}
    \label{fig:Val-CNN}
\end{figure}

Figures~\ref{fig:Val-AttCNN} and \ref{fig:Val-CNN} illustrate the actual (x-axis) vs validated (y-axis) set data for both CNN and its 2-head-attention-augmented counterpart. Both plots show close alignments between the actual and predicted values, indicating that either one would be good enough for the prediction of our leeway objects. 

\begin{figure}[h]
    \centering
 \includegraphics[width=1\linewidth]{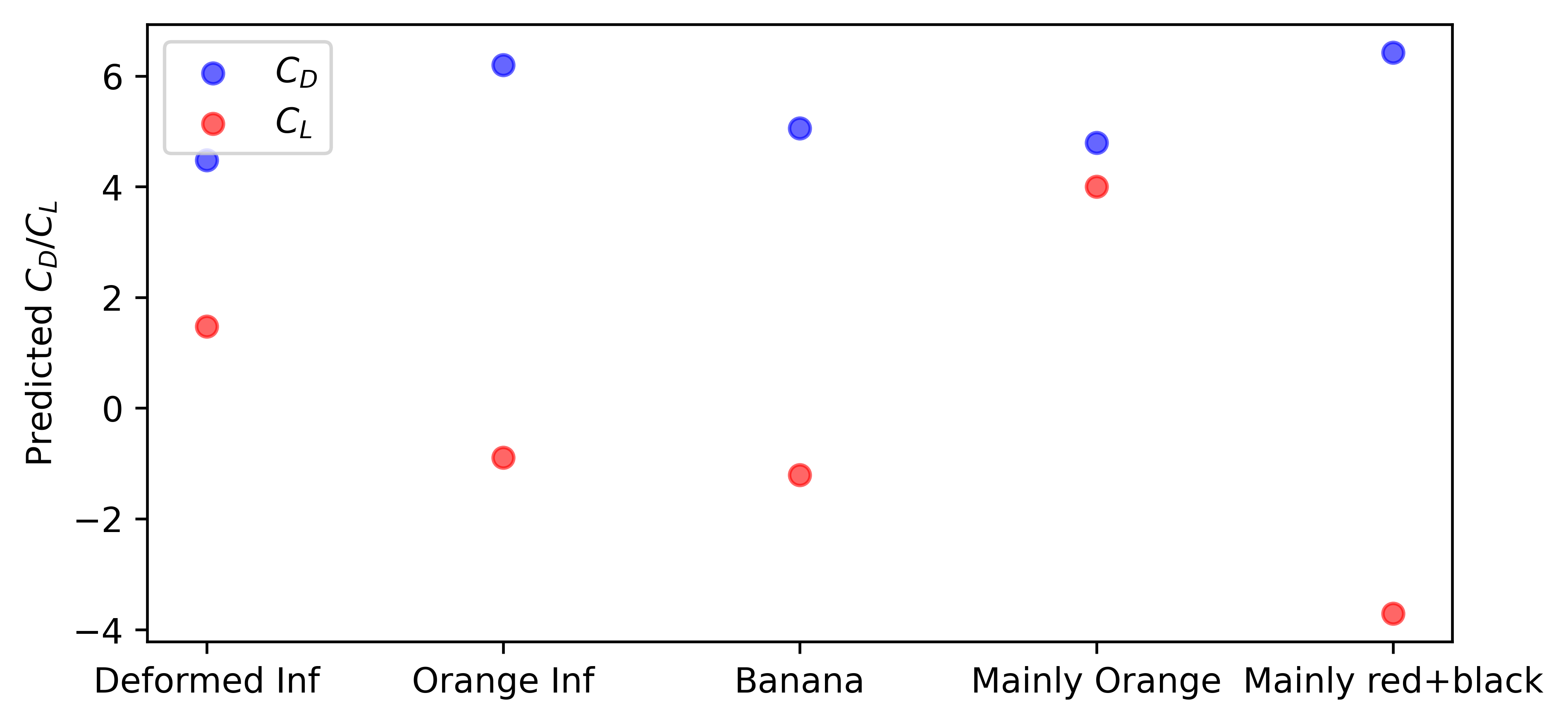}
     \vspace{-0.3in}
    \caption{Estimated Test $C_D$ and $C_L$ of Leeway Objects}
    \label{fig:boat-pred}
\end{figure}

The leeway objects' drag and lift coefficients are estimated based on parameters of the trained lightweight CNN model, because the model achieved the lowest validation scores. Figure~\ref{fig:boat-pred} shows the resulting coefficient values of each boat, used to estimate the object forces reported in Appendix~\ref{App:A}.

\subsection{Drift Prediction Results}
Each leeway object initially has a collated and estimated dataset with a time duration of $0-1500s$. For generalization, data is stacked time-wise to achieve a dataset of total length $7505$. First, we sample data by time-horizon, then to evaluate each object, we train models on the rest of the $4$ object data and $50\%$ of the $5^\mathrm{th}$, and we then test on the remaining $50\%$. We allow up to $1000$ epochs, though typical convergence occurs much less than that time. To mitigate overfitting, the input data without the Sentence Transformer was augmented with synthetic noise. Specifically, a uniform multiplicative noise of $\pm 5\%$ was applied to the original dataset that was combined with the original data, doubling the number of samples. To assess the results of our multi-modal models, MM-Attention-STS-LSTM and MM-STS-Transformer,  we compare their performances against several traditional methods: a recurrent neural network (RNN), temporal convolutional neural network (TCN), sequence-to-sequence LSTM (STS-LSTM), and a classical physics‐based leeway model (Curve Fit) modified from \cite{HowWind}. At steady‐state conditions, the physics‐based leeway model reads $v = v_w +\eta v_a$, for leeway factor $\eta$. However, since directly differentiating noisy velocity time‐series both amplifies measurement error and neglects unsteady wave effects, we simplify the equation into an integral form and add a linear term to account for wave‐induced motion:
\begin{align}
   d^x(t) &= c_{1,x} \int_{0}^{t} v_{w,x}(\tau) \mathrm{d}\tau
          + c_{2,x} \int_{0}^{t} v_{a,x}(\tau) \mathrm{d}\tau
          + c_{3,x} t,\nonumber \\
   d^y(t) &= c_{1,y} \int_{0}^{t} v_{w,y}(\tau) \mathrm{d}\tau
          + c_{2,y} \int_{0}^{t} v_{a,y}(\tau) \mathrm{d}\tau
          + c_{3,y} t,
\end{align}
where each $c_{i,j}$ is a dimensionless coefficient in the $j\in\{x,y\}$ direction.  These coefficients are estimated by fitting the data via nonlinear least‐squares regression. Physically, $c_{1,j}$ quantifies the effect of water currents on the submerged portion of the object, $c_{2,j}$ captures the wind‐induced effect, and $c_{3,j}$ represents the contribution of unsteady wave forcing. The baseline RNN is a two-layer network that takes an input sequence of previous times $\ell_e$ and returns the next time-step output. It consists of a $2$-layer RNN followed by a dense layer to map the outputs to the output dimension. The TCN takes the same input sequence as the RNN to output the next time step. It uses causal convolutions to ensure no future information leaks into the present, maintaining strict temporal integrity. The TCN block is configured without skip connections, uses batch normalization, and applies $\mathrm{ReLU}$ activation with He normal initialization. STS-LSTM is similar to MM-Attention-STS-LSTM without the attention layer and the Sentence Transformer. It takes $10$ past input sequences, similar to RNN and TCN, and STS-LSTM; however, the decoder predicts $10$ future output sequences. We optimize all five ML models using the Adam optimizer with learning rates and other hyperparameters reported in Table~\ref{tab:hyper}, training with MSE loss and evaluating with RMSE, MAE, and MAPE metrics.

\begin{table}[h]
\caption{Hyperparameters of ML Models Reported in this Study}\label{tab:hyper}
\centering
\resizebox{1\linewidth}{!}{
\begin{tabular}{|l|c|c|c|c|c|c|} \hline 
Model& Batch& Learning  & $\ell_e$ &$\ell_d$&Attention & Layer \\ 
& size&  rate &  &&heads& units \\ \hline \hline
\multirow{2}{1.5cm}{RNN} & \multirow{2}{*}{$64$}& \multirow{2}{*}{$1e-3$}& \multirow{2}{*}{$10$} & \multirow{2}{*}{-}& \multirow{2}{*}{-}  & RNN units : $128$, $64$\\
&&&&&&Dense layer : $2$\\ \hline 

\multirow{4}{1.5cm}{TCN} &\multirow{4}{*}{$64$} & \multirow{4}{*}{$1e-3$} & \multirow{4}{*}{$10$}& \multirow{4}{*}{-} & \multirow{4}{*}{-} & Dilations : $32, 16, 8$\\
&&&&&&Kernel size : $11$\\
&&&&&&Activation : $\mathrm{ReLU}$\\
&&&&&&Dense layer :  $2$\\ \hline 

\multirow{3}{1.5cm}{STS-LSTM} & \multirow{3}{*}{$32$}&  \multirow{3}{*}{$1e-3$}&  \multirow{3}{*}{$10$}&  \multirow{3}{*}{$10$}& \multirow{3}{*}{-} &Encoder units : $64, 64$\\ 
&&&&&&Decoder units : $64, 64$\\ 
&&&&&& Dense layer : $2$\\ \hline

\multirow{3}{1.5cm}{MM-Attention-STS-LSTM} & \multirow{3}{*}{$32,16$}  & \multirow{3}{*}{$1e-3$} & \multirow{3}{*}{$10$}&  \multirow{3}{*}{$10$} &  \multirow{3}{*}{$1$} &Encoder units : $64, 64$\\ 
&&&&&&Decoder units : $64$\\ 
&&&&&& Dense layer : $2$\\ \hline 

\multirow{5}{1.5cm}{MM-STS-Transformer}  & \multirow{5}{*}{$32$}  & \multirow{5}{*}{$1e-3$} & \multirow{5}{*}{$10$}&  \multirow{5}{*}{$10$} &  \multirow{2.5}{*}{Encoder : $4$} &$d_m$ : $64$\\ 
&&&&&&$f$ : $128$\\ 
&&&&&\multirow{1.5}{*}{Decoder : $4$}&$d_k$ : $64$\\
&&&&&&Activation : $\mathrm{ReLU}$\\
&&&&&& Dense layer : $2$\\ \hline 

\end{tabular}}
\end{table}

For the ML models, early stopping (patience $= 30$) is employed. Before splitting into sequences, input data is standardized by subtracting its mean ($\mu_{\mathcal{X}}$) and scaling it by its standard deviation ($\sigma_{\mathcal{X}}$):
\begin{align}
\mathcal{X} = \frac{\mathcal{X}_\mathrm{original}- \mu_{\mathcal{X}}}{\sigma_{\mathcal{X}}}.
\end{align}

Table~\ref{tab:thtime} presents the comparative performance of different models across five leeway objects for four prediction horizons ($t_h = 1, 3, 5, 10s$) using three error metrics: RMSE, MAE, and MAPE.
 
 \begin{table*}[!ht]
    \caption{Comparison Results of our Model with Others on Each Boat (best scores in \textcolor{blue}{blue}, second-best scores in \textcolor{cyan}{cyan})}\label{tab:thtime}
    \centering
    \resizebox{1\linewidth}{!}{
\begin{tabular}{|l|l|ccc|ccc|ccc|ccc|}
\hline
Leeway Object & Model & \multicolumn{3}{|c|}{$t_h = 1s$} & \multicolumn{3}{|c|}{$t_h = 3s$} & \multicolumn{3}{|c|}{$t_h = 5s$} & \multicolumn{3}{|c|}{$t_h = 10s$}  \\ 
& & RMSE & MAE & MAPE & RMSE & MAE & MAPE & RMSE & MAE & MAPE & RMSE & MAE & MAPE\\ \hline  \hline
\multirow{6}{1.5cm}{Deformed Inflatable} & Curve fit&43.4547 & 32.2401 & 19.1449 & 43.5543 & 32.3079 & 19.1550 & 44.4454 & 32.9966 & 19.5467 & 44.6538 & 33.1591 & 19.5979\\
& RNN&2.0974 & 1.4970 & 1.8561 & 2.8499 & 1.9181 & 1.8042 & 3.9827 & 2.8126 & 3.0844 & 3.7162 & 2.7581 & 4.0983\\
& TCN&1.8117 & 1.3101 & 1.2406 & 2.5445 & 2.6439 & 1.8646 & 3.0794 & 3.0677 & 2.3176 & 3.4206 & 4.3539 & 2.6839\\
& STS-LSTM&1.6057 & 1.0024 & {\color{blue}0.8807} & 2.7203 & 1.6361 & 1.5733 & 2.7203 & 1.6361 & {\color{cyan}1.5733} & 3.6361 & {\color{cyan}1.9340} & {\color{cyan}1.8486}\\
& MM-Attention-STS-LSTM&{\color{cyan}1.3850} & {\color{blue}0.9747} & {\color{cyan}1.1715} & {\color{cyan}1.7925} & {\color{cyan}1.2329} & {\color{cyan}1.4066} & {\color{cyan}2.2534} & {\color{cyan}1.5631} & 1.6745 & {\color{cyan}2.7429} & 2.0037 & 2.1716\\
& MM-STS-Transformer&{\color{blue}1.2803} & {\color{cyan}1.0008} & 1.3560 & {\color{blue}1.4762} & {\color{blue}1.0476} & {\color{blue}1.0459} & {\color{blue}1.5729} & {\color{blue}1.1403} & {\color{blue}1.1733} & {\color{blue}1.7560} & {\color{blue}1.2616} & {\color{blue}1.4927}\\ \hline

\multirow{6}{1.5cm}{Orange Inflatable} & Curve fit&56.4812 & 41.7997 & 37.0432 & 57.1196 & 42.2527 & 37.1991 & 56.7867 & 42.0365 & 37.2664 & 55.3529 & 41.0452 & 37.1208\\
& RNN&1.3747 & 1.1255 & 4.7596 & 91.9490 & 63.9734 & 43.3703 & 1.8270 & 1.4029 & 4.9129 & {\color{cyan}1.9244} & {\color{cyan}1.3946} & 4.1710\\
& TCN&{\color{cyan}0.8630} & 2.7853 & {\color{blue}0.7082} & {\color{blue}1.1092} & 2.5785 & {\color{blue}0.8162} & 1.5209 & 3.0101 & {\color{blue}1.1217} & 2.4956 & 4.4905 & {\color{blue}1.7644}\\
& STS-LSTM&0.8896 & {\color{cyan}0.6356} & {\color{cyan}1.6789} & 1.2809 & {\color{cyan}0.8596} & 2.3800 & 1.6658 & 1.1754 & 2.9980 & 2.0687 & 1.4632 & 4.7713\\
& MM-Attention-STS-LSTM&0.9668 & 0.6755 & 1.8048 & {\color{cyan}1.1951} & {\color{blue}0.7909} & {\color{cyan}1.8496} & {\color{cyan}1.1234} & {\color{blue}0.8009} & {\color{cyan}2.2089} & 2.1024 & 1.5562 & 4.0854\\
& MM-STS-Transformer&{\color{blue}0.7501} & {\color{blue}0.5571} & 1.7078 & 1.2363 & 0.8944 & 2.5203 & {\color{blue}1.0172} & {\color{cyan}0.8042} & 2.7342 & {\color{blue}1.3973} & {\color{blue}1.0208} & {\color{cyan}2.6777}\\
\hline

\multirow{6}{1.5cm}{Banana} & Curve fit&32.1320 & 23.9958 & 13.3140 & 32.0867 & 23.9679 & 13.2892 & 32.2965 & 24.1221 & 13.3599 & 31.8891 & 23.8335 & 13.1801\\
& RNN&0.8128 & 0.6001 & 0.4553 & 1.8486 & 1.3434 & 0.9604 & 1.9091 & 1.4127 & 1.0382 & 5.5431 & 3.6948 & 2.8617\\
& TCN&{\color{cyan}0.7286} & {\color{blue}0.3958} & 0.5601 & {\color{cyan}1.0581} & {\color{blue}0.6139} & 0.8461 & {\color{blue}1.1266} & {\color{blue}0.7149} & 0.8893 & {\color{cyan}1.4571} & {\color{blue}0.7682} & {\color{cyan}1.0942}\\
& STS-LSTM&0.8189 & 0.6015 & {\color{cyan}0.4215} & 1.5741 & 1.1321 & 0.8300 & 1.5576 & 1.0961 & {\color{cyan}0.8224} & 2.5751 & 1.7009 & 1.1123\\
& MM-Attention-STS-LSTM&0.9033 & 0.6715 & 0.5119 & 1.1282 & 0.7746 & {\color{blue}0.5236} & 1.7128 & 1.1844 & 0.8554 & 2.9529 & 1.9495 & 1.3484\\
& MM-STS-Transformer&{\color{blue}0.6095} & {\color{cyan}0.4035} & {\color{blue}0.3184} & {\color{blue}0.9859} & {\color{cyan}0.7250} & {\color{cyan}0.5415} & {\color{cyan}1.1770} & {\color{cyan}0.8231} & {\color{blue}0.5837} & {\color{blue}1.4070} & {\color{cyan}1.0016} & {\color{blue}0.6755}\\
\hline

\multirow{6}{1.5cm}{Mainly orange 3-D printed} & Curve fit&29.4190 & 20.9228 & 12.8642 & 29.3459 & 20.8663 & 12.8191 & 29.4450 & 20.9349 & 12.8376 & 29.3305 & 20.8443 & 12.7539\\
& RNN&1.7312 & 1.4242 & 4.5936 & 1.4701 & 1.1440 & 2.7827 & 4.6715 & 3.2087 & 5.9435 & 1.8103 & 1.4487 & 4.2234\\
& TCN&{\color{cyan}0.6658} & 1.1239 & {\color{blue}0.5039} & 1.2683 & 1.7003 & {\color{blue}0.9605} & 1.4223 & 2.4876 & {\color{blue}1.0835} & 1.7781 & 2.4620 & {\color{cyan}1.3316}\\
& STS-LSTM&0.7620 & {\color{cyan}0.5317} & {\color{cyan}0.9153} & 1.2611 & {\color{cyan}0.8212} & {\color{cyan}1.0147} & {\color{cyan}1.0444} & {\color{cyan}0.7037} & {\color{cyan}1.1257} & {\color{blue}0.9492} & {\color{blue}0.7473} & 1.6977\\
& MM-Attention-STS-LSTM&0.7698 & 0.5747 & 1.1891 & {\color{cyan}1.1129} & 0.8631 & 2.0436 & 1.0876 & 0.8439 & 1.8344 & 2.0794 & 1.4884 & 3.0352\\
& MM-STS-Transformer&{\color{blue}0.6475} & {\color{blue}0.4944} & 1.0551 & {\color{blue}0.7937} & {\color{blue}0.5872} & 1.1485 & {\color{blue}0.8631} & {\color{blue}0.6735} & 1.6127 & {\color{cyan}1.3877} & {\color{cyan}0.8794} & {\color{blue}1.3204}\\ \hline

\multirow{6}{1.5cm}{Mainly red+black 3-D printed} & Curve fit&36.9403 & 27.0266 & 21.8883 & 36.9122 & 27.0109 & 21.8764 & 36.9014 & 27.0077 & 21.8736 & 36.8037 & 26.9482 & 21.8264\\
& RNN&0.9630 & 0.7853 & 1.7437 & 1.4477 & 1.1631 & 2.3420 & 1.4634 & 1.1779 & 2.5780 & 1.8119 & 1.4764 & 3.5498\\
& TCN&{\color{cyan}0.5771} & 0.6920 & {\color{blue}0.4247} & 1.0713 & 1.4282 & {\color{blue}0.8079} & 1.4291 & 2.0175 & 1.0897 & 1.4349 & 1.6317 & {\color{cyan}1.0264}\\
& STS-LSTM&0.6055 & {\color{cyan}0.4153} & {\color{cyan}0.7002} & 0.9342 & 0.6367 & 0.9424 & {\color{cyan}0.7642} & {\color{cyan}0.5467} & {\color{cyan}0.9884} & {\color{blue}0.8958} & {\color{blue}0.6227} & 1.0579\\
& MM-Attention-STS-LSTM&0.6832 & 0.5003 & 0.9149 & {\color{cyan}0.8197} & {\color{cyan}0.6020} & 1.1409 & 0.9500 & 0.6856 & 1.1203 & 1.3108 & 0.9322 & 1.5595\\
& MM-STS-Transformer &{\color{blue}0.3949} & {\color{blue}0.3184} & 0.7010 & {\color{blue}0.5945} & {\color{blue}0.4396} & {\color{cyan}0.8132} & {\color{blue}0.6954} & {\color{blue}0.5065} & {\color{blue}0.8832} & {\color{cyan}0.9956} & {\color{cyan}0.7254} & {\color{blue}1.0043}\\ \hline
\end{tabular}}
\end{table*}

For each metric and horizon, the two lowest errors are highlighted (best in blue, second-best in cyan. Across almost all leeway objects and time horizons, the MM-STS-Transformer consistently achieves the lowest or second-lowest errors in all three metrics, indicating its ability to capture both temporal dependencies and textual relationships induced by multi-modal dynamics, valuable for drift prediction. The MM-Attention-STS-LSTM or STS-LSTM are often the second-best performers, suggesting that attention-augmented recurrent models remain competitive. At short horizons, errors are significantly smaller overall, and the ranking gap between top models and baselines (Curve Fit, RNN) is more pronounced. All models experience increasing RMSE, MAE, and MAPE as $t_h$ grows from $1s$ to $10s$; however, the MM-STS-Transformer exhibits a slower degradation rate, especially for RMSE, indicating better long-term generalization. Performance gains from the transformer approach are most notable in objects like the Deformed Inflatable and  Banana, conversely, more streamlined shapes (e.g., Mainly orange 3-D printed) show smaller performance gaps between advanced models and simpler baselines. Curve Fit and plain RNN models produce errors an order of magnitude larger than deep learning models, further reinforcing the importance of nonlinear feature extraction and sequence modeling in drift prediction.

\begin{figure}[!ht]
    \centering
    \includegraphics[width = \linewidth]{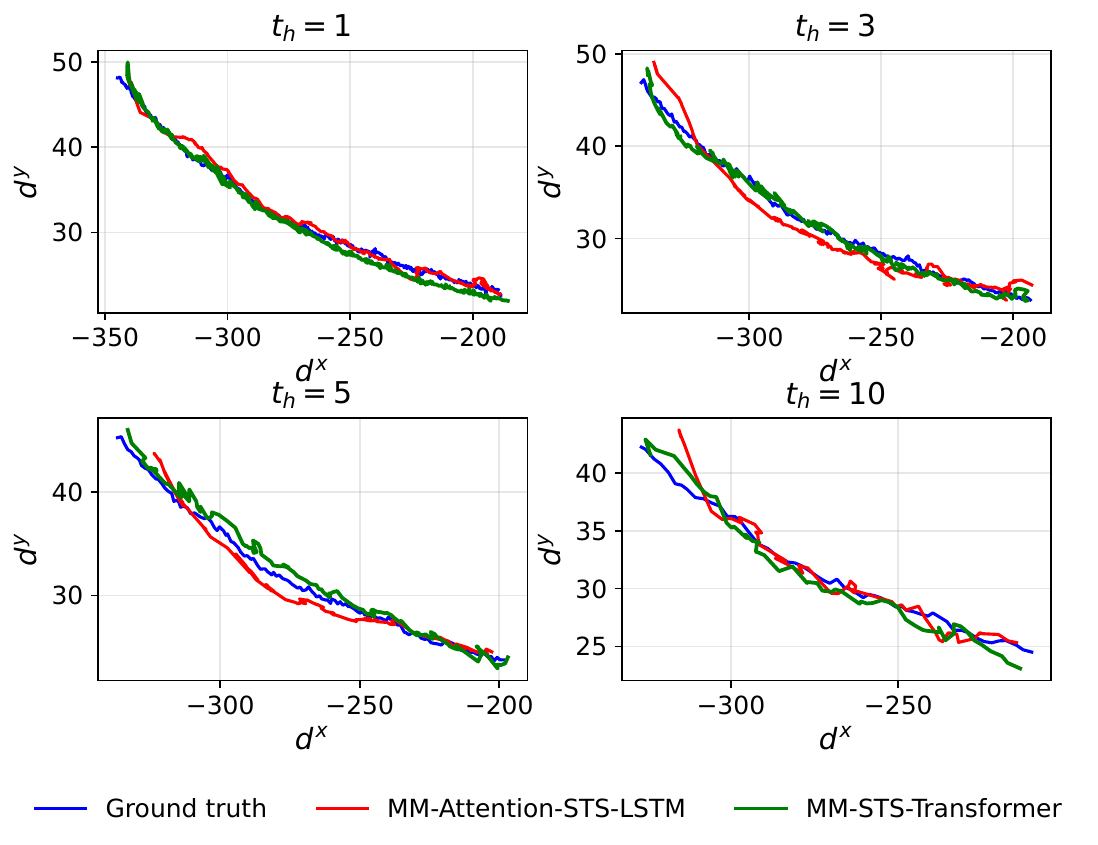}
     \vspace{-0.3in}
   \caption{Comparison plots of the actual drift trajectory for the deformed inflatable boat with time horizon of  $1s$ (upper left), $3s$ (upper right), $5s$ (lower left) and $10s$ (lower right) } \label{fig:B1results}
\end{figure}

\begin{figure}[!ht]
    \centering
     \includegraphics[width = \linewidth]{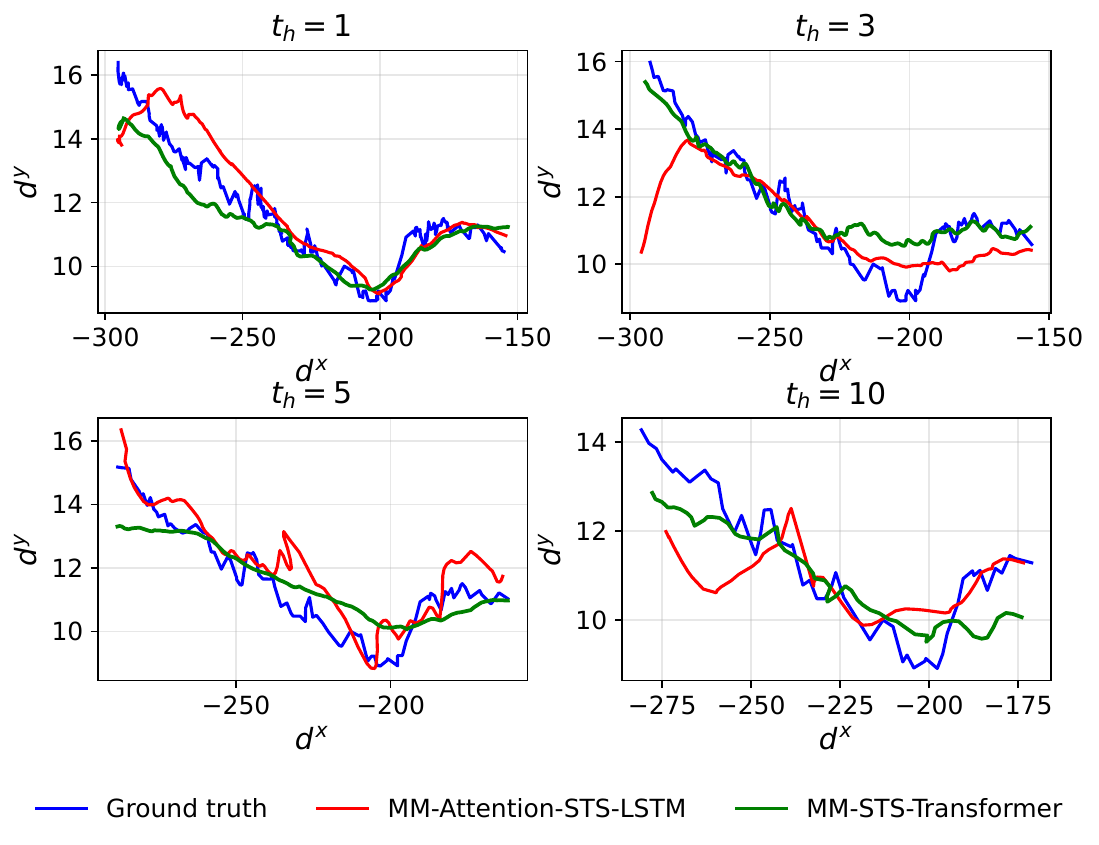}
     \vspace{-0.3in}
   \caption{Comparison plots of the actual drift trajectory for the orange inflatable boat with time horizon of  $1s$ (upper left), $3s$ (upper right), $5s$ (lower left) and $10s$ (lower right) } \label{fig:B2results}
\end{figure}

\begin{figure}[!t]
    \centering
     \includegraphics[width = \linewidth]{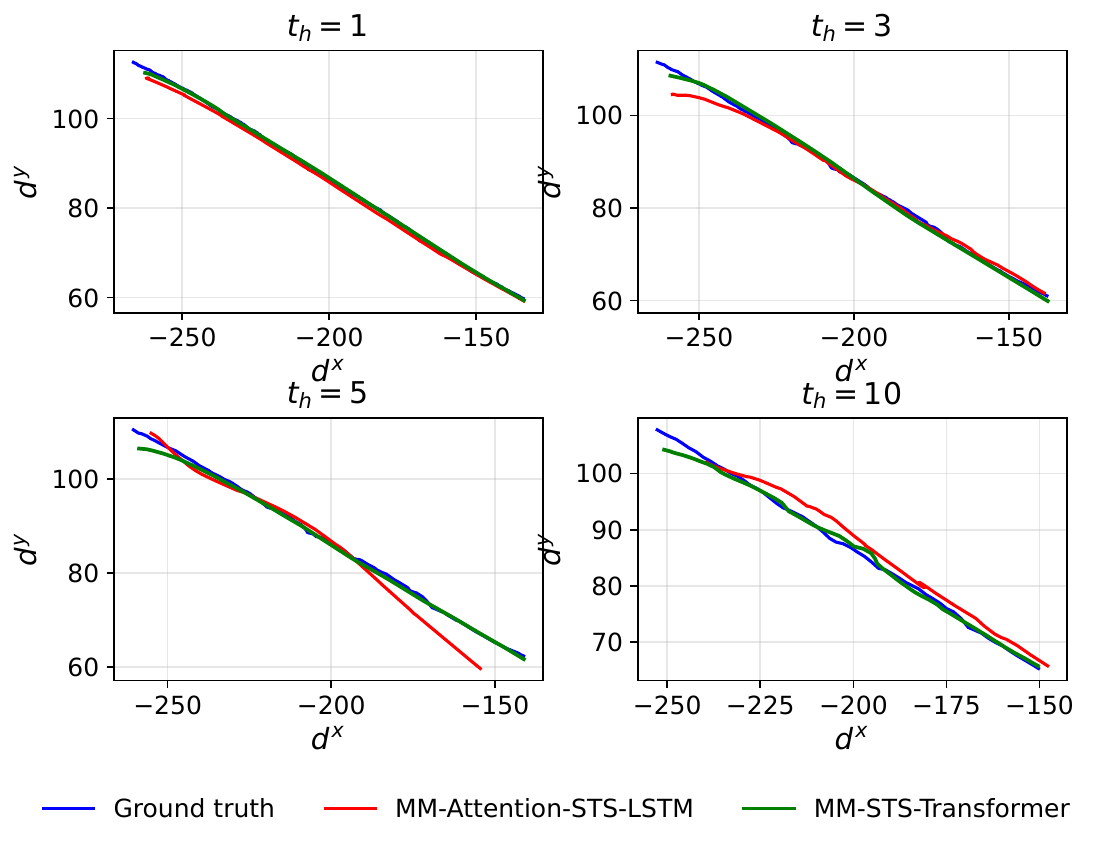}
     \vspace{-0.3in}
   \caption{Comparison plots of the actual drift trajectory for the banana boat with time horizon of  $1s$ (upper left), $3s$ (upper right), $5s$ (lower left) and $10s$ (lower right) } \label{fig:B3results}
\end{figure}

\begin{figure}[!ht]
    \centering
   \includegraphics[width = \linewidth]{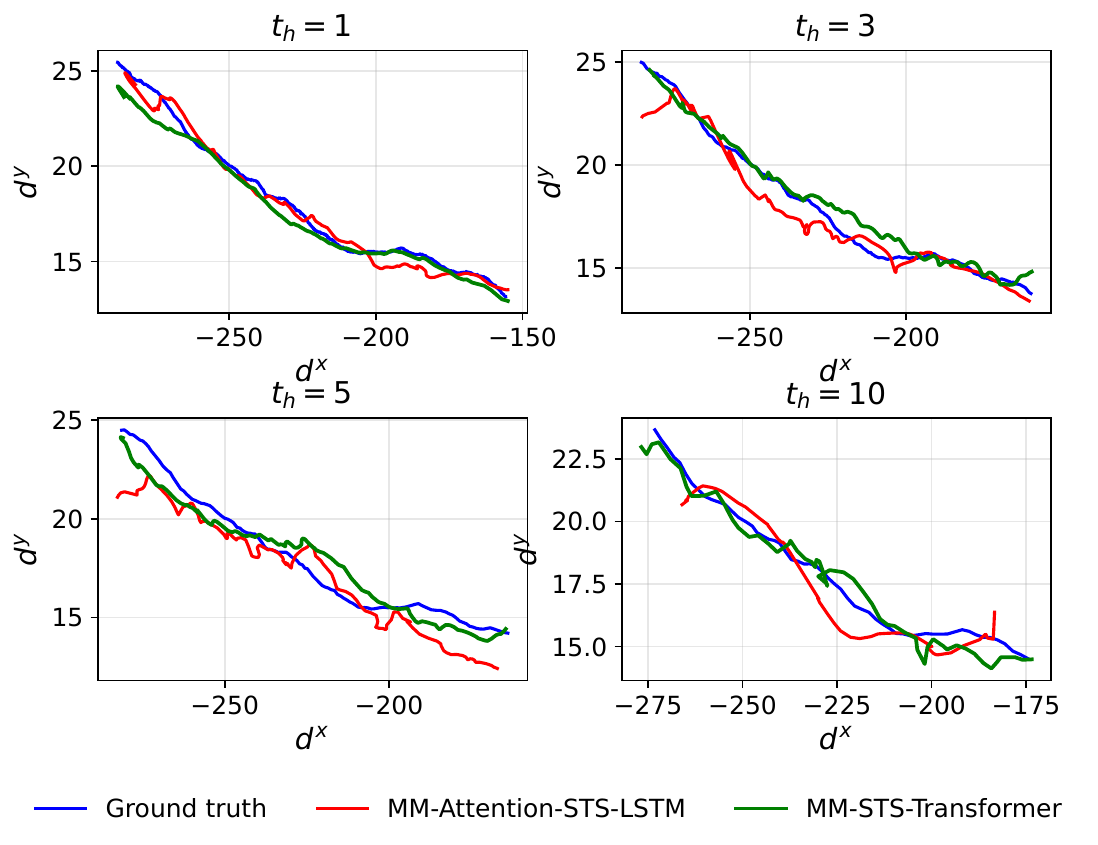}
     \vspace{-0.3in}
   \caption{Comparison plots of the actual drift trajectory for the mainly orange 3-D printed boat with time horizon of  $1s$ (upper left), $3s$ (upper right), $5s$ (lower left) and $10s$ (lower right) } \label{fig:B4results}
\end{figure}

\begin{figure}[!ht]
    \centering
    \includegraphics[width = \linewidth]{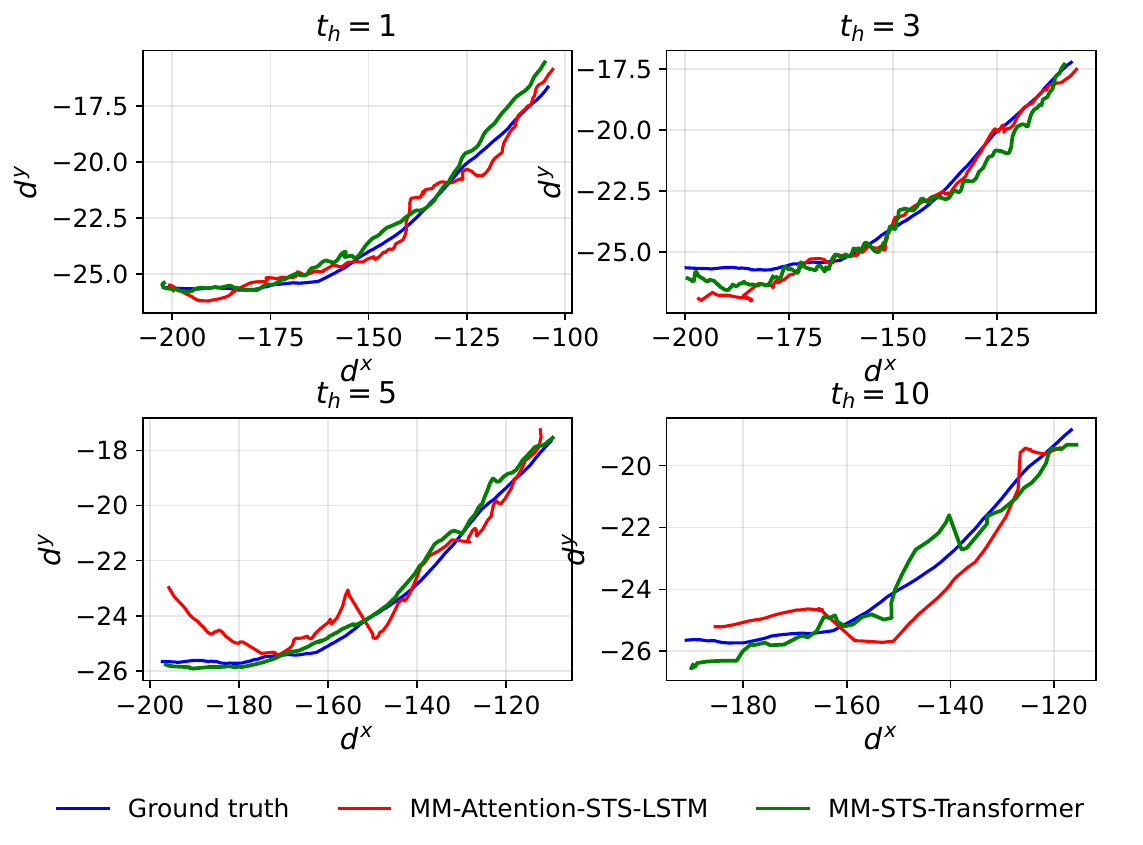}
     \vspace{-0.3in}
   \caption{Comparison plots of the actual drift trajectory for the mainly red+black 3-D printed boat with time horizon of  $1s$ (upper left), $3s$ (upper right), $5s$ (lower left) and $10s$ (lower right) } \label{fig:B5results}
\end{figure}
 Qualitative trajectory overlays (included in the Figures~ [\ref{fig:B1results}, \ref{fig:B2results}, \ref{fig:B3results}, \ref{fig:B4results}, \ref{fig:B5results}] corroborate the trends discussed. The figures show the test data trajectory plot of the ground truth against the MM-Attention-STS-LSTM and MM-STS-Transformer prediction results at each $t_h$ sampling, indicating that the MM-STS-Transformer maintains a tighter spread over MM-Attention-STS-LSTM.

\section{Discussions}~\label{Sec:Discussions}

\paragraph{Interpretation of Drag and Lift Estimation Results}
In general, the lightweight CNN achieved the lowest MSE and MAE, likely due to its reduced parameter count, mitigating overfitting on the relatively small dataset ($179$ images). Attention-augmented CNNs, while theoretically better at capturing global dependencies, added complexity that the dataset size could not fully exploit. $2$-head-attention improved over single-head setups but still underperformed the lightweight counterpart. 

\paragraph{Interpretation of Drift Prediction Results}
Across nearly all objects and horizons, MM-STS-Transformer yielded the lowest or second-lowest RMSE, MAE, and MAPE. This likely stems from the model's attention layers and multi-modal fusion (temporal and textual) of the input data, enabling better long-horizon generalization. In the case of the MM-Attention-STS-LSTM, it is a sequence-to-sequence model predicting $ 10 $ times into the future. Compared to RNN and Curve fit results, which forecast just one time step ahead, its resulting scores are comparably efficient. In general, far-ahead forecasting is most especially good in SAR operations, as it translates to faster response time. The physics-based curve fit model shows a lack of generalizability across boats and would work best when coefficients are individually obtained for each boat. In addition to these results, since experiments were performed at the same time, the water and wind velocities are equal for all boats, and so using only these parameters as input alone would translate to poor generalization for the model. The higher accuracies in the other models, compared to the curve fit, can be related to the additional inputs, such as the forces and the textual descriptions. Despite this, all models degrade as the prediction horizon increases, but the transformer model decays the slowest, making it the best-performing model overall. Irregular shapes (Deformed Inflatable, Banana) benefit most from deep multi-modal architectures, while streamlined shapes (Mainly orange 3-D printed) show smaller performance gaps between advanced and simpler models.

\paragraph{Practical Implications and Limitations}
The practical implications of our models are in the improved drift forecasts that are capable of reducing the required search area and response time in SAR operations. The same framework could support tracking pollutants or floating debris by adapting the object-specific input modality. 
While this research has shown the feasibility of multi-modal ML modeling of drift forecasting, one major experiment limitation it has is that it was performed in a confined lake where the aerodynamic and hydrodynamic parameters are almost constant. There is an experimental time window constraint. The object drifts more slowly than in the actual sea, and the rate of movement is very low. In addition to this, experiments were limited to five objects under specific conditions; generalization to other shapes, materials, or environments is not yet validated. Finally, in this study, we have assumed that the air and lift coefficients for both air and water are equal, which is likely untrue in the real world. These coefficients are dependent on the Reynolds number and the velocity of the fluid in question. The lift and drag coefficients modeled in this study are technically related to those of air, and later works may improve on these by building simulation models for both air and water coefficients.

\section{Conclusion}\label{Sec:Conclusions}
In this study, we presented a complete multi-modal framework that integrates a Language Model, in the form of a Sentence Transformer, with attention-based sequence-to-sequence architectures to improve drift forecasting for leeway objects in water. We experimentally observed and collated key environmental parameters for five distinct leeway objects. We trained a lightweight CNN on image data obtained from a Navier-Stokes-based simulation model to estimate drag and lift coefficients of the five objects. These coefficients were then used to derive the net forces driving object motion, which are the temporal data for our models. Temporal data, along with object-specific textual descriptions, were input into our proposed multi-modal attention-based sequence-to-sequence prediction models: Attention-based sequence-to-sequence LSTM (MM-Attention-STS-LSTM) and sequence-to-sequence Transformer (MM-STS-Transformer). The models predict drift trajectories over time horizons of $1$, $3$, $5$, and $10$ seconds. We compare results against a curve-fitted physics model and traditional machine learning baselines, demonstrating that the MM-STS-Transformer achieved the highest accuracy, particularly for irregularly shaped objects and longer prediction horizons, and MM-Attention-STS-LSTM remains a strong contender, outperforming models that predict one time step ahead, such as the curve fit and the RNN models. The results confirm that attention-based sequence-to-sequence incorporating multi-modal fusion offers robust and generalizable performance. Findings highlight the value of integrating physically relevant parameters, textual descriptions with deep sequence modeling for operational applications in search and rescue and marine environmental monitoring. Future work may focus on expanding the dataset to include diverse object geometries, materials, and environmental conditions; integrating high-resolution wave, water, and current field data; and exploring physics-informed transformer architectures to embed constraints to capture robust real-world settings.
\nomenclature{$\mathcal{X}$}{Multi‑modal input data}
\nomenclature{$C_D^a$,\;$C_D^w$}{Drag coefficient for air/water flow}
\nomenclature{$C_L^a$,\;$C_L^w$}{Lift coefficient for air/water flow}
\nomenclature{$D_a$,\;$D_w$}{Drag force magnitude from air/water [$\mathrm{N}$]}
\nomenclature{$D_{a,x},D_{a,y}$}{Eastward/Northward components of  [$\mathrm{N}$]}
\nomenclature{$D_{w,x},D_{w,y}$}{Eastward/Northward components of water drag force [$\mathrm{N}$]}
\nomenclature{$\rho_a$,\;$\rho_w$}{Density of air/water ($1.225 \mathrm{kg/m}^3$/$1025 \mathrm{kg/m}^3$ )}
\nomenclature{$v$}{Velocity vector of the drifting object  [$\mathrm{m/s}$]}
\nomenclature{$v_w$,\;$v_a$ }{Water current /air velocity $\mathrm{m/s}$}
\nomenclature{$t_h$}{Time horizon}
\nomenclature{$\mathcal{Y}= \{d^x, d^y\}$}{Eastward/Northward drift coordinates (output) [$\mathrm{m}$]}
\nomenclature{$v_{a,x},v_{a,y}$}{Eastward/Northward components of $v_a$ [$\mathrm{m/s}$]}
\nomenclature{$v_{w,x},\,v_{w,y}$}{Eastward/Northward components of $v_w$ [$\mathrm{m/s}$]}
\nomenclature{$\tilde v_{a,x},\tilde v_{a,y}$}{Eastward/Northward components of $\tilde v_a$ [$\mathrm{m/s}$]}
\nomenclature{$\tilde v_{w,x},\tilde v_{w,y}$}{Eastward/Northward components of $\tilde v_w$ [$\mathrm{m/s}$]}
\nomenclature{$A_a$,\;$A_w$}{Surface area exposed to wind/water loading [$\mathrm{m}^2$]}
\nomenclature{$m_o$}{Mass of drifting object [$\mathrm{kg]}$}
\nomenclature{$v_{\mathrm{ref}}$}{Measured wind speed at reference height $z_{\mathrm{ref}}$ [$\mathrm{m/s}$]}
\nomenclature{$z$}{Target height for wind extrapolation $(10\,m)$ [$\mathrm{m}$]}
\nomenclature{$z_{\mathrm{ref}}$}{Reference height of wind measurement $(2.0066)$ [$\mathrm{m}$]}
\nomenclature{$v_x,v_y$}{Eastward/Northward of $v$ [$\mathrm{m/s}$]}
\nomenclature{$\tilde v_a$\;$\tilde v_w$}{Relative wind/water  current speed [$\mathrm{m/s}$]}
\nomenclature{$T$}{Total number of time steps}
\nomenclature{$\beta$}{Exponent in power‑law wind profile}
\nomenclature{$L_{a,x},L_{a,y}$}{Eastward/northward air‑lift components [$\mathrm{N}$]}
\nomenclature{$L_{w,x},L_{w,y}$}{Eastward/northward water‑lift components [$\mathrm{N}$]}
\nomenclature{$L_a$\; $L_w$}{Lift force magnitude from air/water [$\mathrm{N}$]}
\nomenclature{$\gamma$}{Submersion rate of boat}
\nomenclature{$\ell_e$}{Encoder window length (timesteps)}
\nomenclature{$\ell_d$}{Decoder (prediction) horizon (timesteps)}

\nomenclature{$p$}{Dimensionality of input data}
\nomenclature{$N_c$}{Number of valid training examples in CNN}

\nomenclature{$N_d$}{Number of valid training examples in LSTM and Transformer}
\nomenclature{$\mathcal{X}^e_i$}{Encoder input slice $\in\mathbb{R}^{\ell_e\times p}$}
\nomenclature{$\mathcal{Y}^d_i$}{Decoder input with teacher forcing $\in\mathbb{R}^{\ell_d\times 2}$}
\nomenclature{$\mathcal{Y}^{\mathrm{out}}_i$}{Decoder target output $\in\mathbb{R}^{\ell_d\times 2}$}

\nomenclature{$\mathbf{e}_i$}{Sentence‐Transformer embedding of object description $d_i$ ($\in\mathbb{R}^{384}$)}
\nomenclature{$I$}{Input image to CNN for drag/lift coefficients prediction}
\nomenclature{$d_m$}{Transformer embedding dimension}
\nomenclature{$H$}{Number of attention heads}
\nomenclature{$d_k$}{Per‑head key/query dimension}
\nomenclature{$f$}{Feed‑forward hidden dimension}

\nomenclature{$\xi$}{Learning rate}
\nomenclature{$\eta$}{Leeway factor}
\printnomenclature

\bibliographystyle{IEEEtran}
\bibliography{ref}
\appendices
\section{Numerical Input Dataset Used in This Study }\label{App:A}

\begin{figure}[!h]
    \centering
    \includegraphics[width=0.7\linewidth]{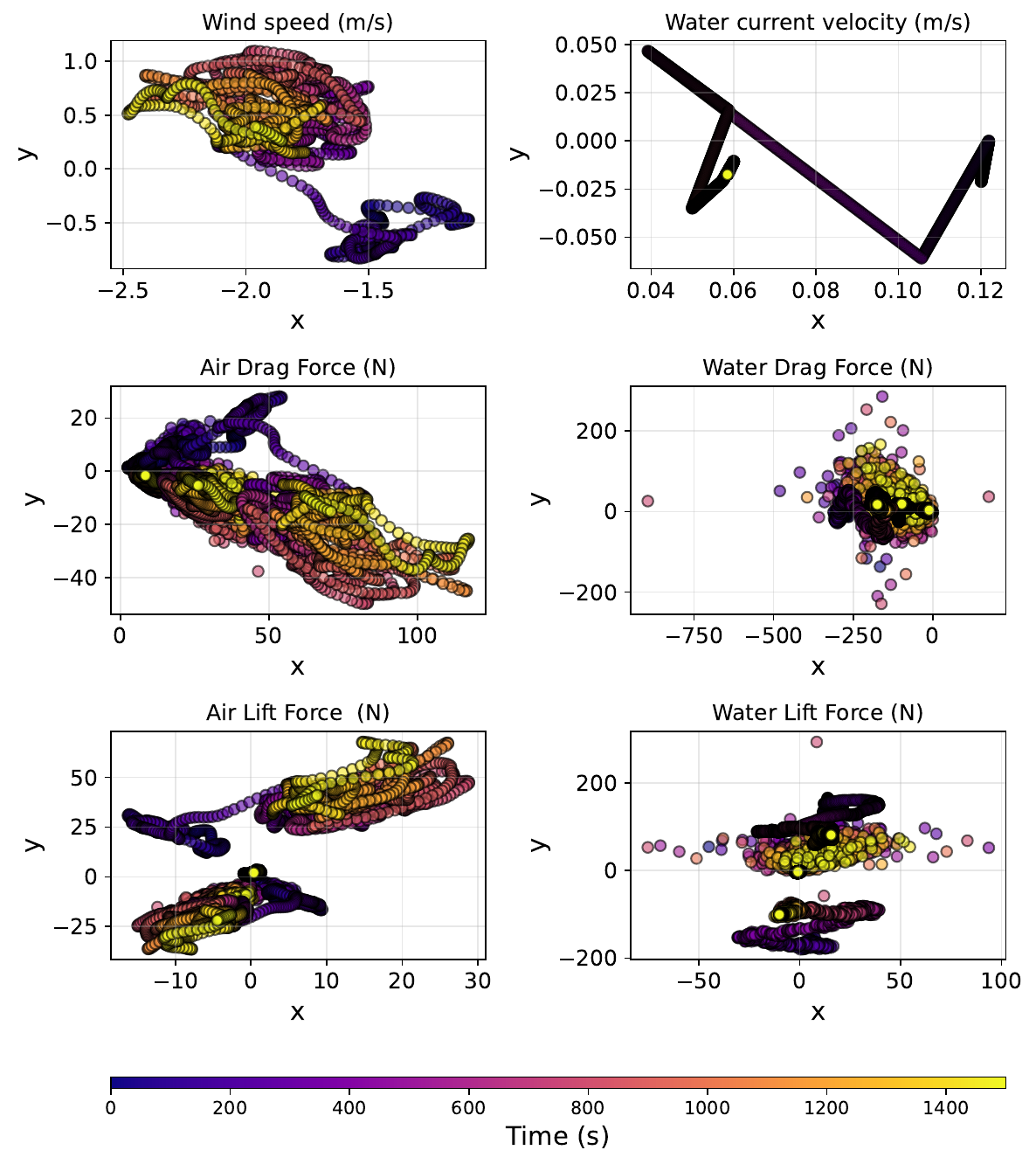}
    \caption{Plots of numerical input dataset used in this study with respect to time}
    \label{fig:enter-label}
\end{figure}

\end{document}